\documentclass{article}

\usepackage[usenames,dvipsnames,svgnames,x11names]{xcolor}

\usepackage[nocompress]{cite}

\usepackage[T1]{fontenc} %% 
\usepackage[normalem]{ulem} % required for strikeout font

\usepackage{dsfont}

\usepackage{balance}

\usepackage{listings}

\lstdefinelanguage{XML}
{
basicstyle=\ttfamily\footnotesize,
  morestring=[b]",
  moredelim=[s][\bfseries\color{Maroon}]{<}{\ },
  moredelim=[s][\bfseries\color{Maroon}]{</}{>},
  moredelim=[l][\bfseries\color{Maroon}]{/>},
  moredelim=[l][\bfseries\color{Maroon}]{>},
  morecomment=[s]{<?}{?>},
  morecomment=[s]{<!--}{-->},
  commentstyle=\color{gray},
  stringstyle=\color{blue},
  identifierstyle=\color{red}
}
\usepackage{moreverb}

\usepackage[nounderscore]{syntax}

\usepackage[pdftex]{graphicx}
\graphicspath{{./figures/}}
\DeclareGraphicsExtensions{.pdf}

\usepackage[cmex10]{amsmath}
\usepackage{amssymb}
\usepackage{mathtools}
\usepackage{amsfonts}

\usepackage{bm}

\usepackage{subfig} %[caption=false,font=footnotesize,labelfont=sf,textfont=sf]
\usepackage{makecell}

\usepackage{algorithmicx}
\usepackage{algpseudocode}
\usepackage[ruled]{algorithm}
\definecolor{light-gray}{gray}{0.75}
\algrenewcommand{\algorithmiccomment}[1]{\hskip3em{{\footnotesize \textcolor{light-gray}{$\blacktriangleright$}}} #1}

\usepackage{multirow} % Multi-row tables
\usepackage{rotating} % sideways table
\usepackage{booktabs} % better lines
\usepackage{colortbl} % cell colors
\usepackage{tablefootnote} % add support for footnote in table

\usepackage{xspace}

\usepackage{enumitem}

\newcommand{\midas}{MiDaS\xspace}
\newcommand{\neo}{\textsc{Neo}\xspace}

\usepackage{blindtext}

\usepackage[pdftex,colorlinks=true,urlcolor=blue,citecolor=blue]{hyperref}
\title{
\LARGE \bf NeoARCADE: Robust Calibration for Distance Estimation to Support Assistive Drones for the Visually Impaired}

\date{}

\begin{document}

\author{Suman Raj, Bhavani A Madhabhavi\thanks{\em were with Dream:Lab at the time of writing this paper.}, Madhav Kumar\thanks{\em Equal Contribution}~\footnotemark[1],\and Prabhav Gupta\footnotemark[2]~\footnotemark[1], and Yogesh Simmhan
\and Department of Computational and Data Sciences,
\and Indian Institute of Science, Bangalore 560012 INDIA
\and Email: \{sumanraj, simmhan\}@iisc.ac.in}

\maketitle

\thispagestyle{plain}
\pagestyle{plain}

\begin{abstract}
    Autonomous navigation by drones using onboard sensors, combined with deep learning and computer vision algorithms, is impacting a number of domains. We examine the use of drones to autonomously follow and assist Visually Impaired People (VIPs) in navigating urban environments. 
    Estimating the absolute distance between the drone and the VIP, and to nearby objects, is essential to design obstacle avoidance algorithms.
    Here, we present NeoARCADE (\neo), which uses depth maps over monocular video feeds, common in consumer drones, to estimate absolute distances to the VIP and obstacles. \neo proposes robust calibration technique based on depth score normalization and coefficient estimations to translate relative distances from depth map to absolute ones. It further develops a dynamic recalibration method that can adapt to changing scenarios. We also develop two baseline models, Regression and Geometric, and compare Neo with SOTA depth map approaches and the baselines. We provide detailed evaluations to validate their robustness and generalizability for distance estimation to VIPs and other obstacles in diverse and dynamic conditions, using datasets collected in a campus environment. \neo predicts distances to VIP with an error $<30cm$, and to different obstacles like cars and bicycles within a maximum error of $60cm$, which are better than the baselines. Neo also clearly out-performs SOTA depth map methods, reporting errors up to $5.3$--$14.6\times$ lower.
\end{abstract}

\section{Introduction }
There is a growing trend for fleets of Unmanned Aerial Vehicles (UAVs), also called drones\footnote{We use the terms UAVs and drones interchangeably in this paper.}, being used for food delivery, flying ambulances~\cite{drone-first-aid}, urban safety monitoring, and disaster response~\cite{ollero2004motion}. UAVs have become popular due to their agility in urban spaces and their lightweight structure. Further, autonomous navigation of UAVs is made possible by the integration of sensors such as GPS, cameras, accelerometers and LIDAR that help the drone perceive and understand its environment~\cite{10.1145/3450356}. While LIDAR and depth cameras are commonly used for Simultaneous Localization and Mapping (SLAM) in UAVs, recent advances in Deep Neural Networks (DNN) models and Computer Vision (CV) algorithms are enabling the use of monocular video streams from onboard cameras, available in even low-end drones, to achieve situation awareness and make autonomous navigation decisions, such as avoiding obstacles or navigating to a destination~\cite{arafat2023vision}. 
Further, advances in GPU edge accelerators and 4G/5G communications make it feasible to perform such real-time inferencing over video streams using onboard accelerators and/or remote cloud servers~\cite{10171496,RAJ2025107874}.

%% #########################################################
\paragraph{Motivating Application}
While there are numerous applications of drones using autonomous navigation, we explore the use of drones as an assistive technology~\cite{al2016exploring}. According to the World Health Organization (WHO) over 2.2B people suffer from moderate or severe visual impairment worldwide, of whom 36M are blind~\cite{whoStats}. This impacts their quality of life and can lead to social isolation, difficulty in mobility, and a higher risk of falls. We examine the use of drones to assist \textit{Visually Impaired People (VIPs)} navigate outdoors, thus helping them lead an active urban lifestyle~\cite{suman2023chi}. 

\begin{figure}[t]
\centering%~%
    \subfloat[Geometric (cyan) and Regression (green) methods]{
    \quad \includegraphics[width=0.4\columnwidth]{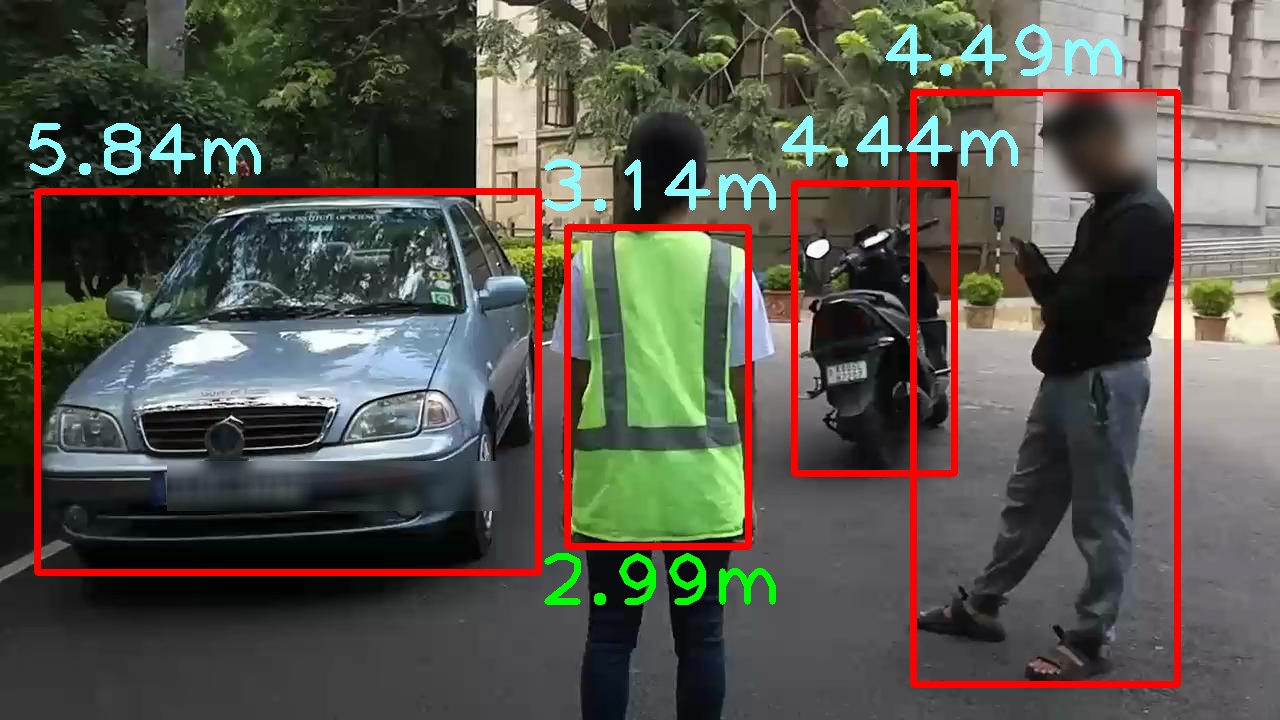} \quad 
   \label{fig:teaser1}
  }\qquad
  \subfloat[\neo method]{
    \includegraphics[width=0.4\columnwidth]{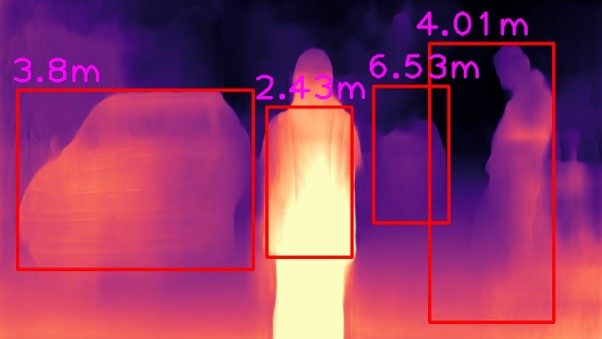}
   \label{fig:teaser2}
  }
\caption{Distance estimation using proposed methods. True distance to VIP is $3m$, bystander is $4m$, car is $4.6m$.}
\label{fig:teaser}
\end{figure}

Among contemporary technologies for navigation assistance of VIPs, smart canes~\cite{wewalk} and smart wearables~\cite{bai2017smart,lookout} offer sensor and video-based guidance but suffer from a restricted range and Field of View (FoV) that limits the detection of hazards. Recent studies are exploring drone-based solutions. Nasralla et al.~\cite{nasralla2019computer} examine the potential of CV-enabled UAVs to aid VIPs in a smart city. 
\textit{DroneNavigator}~\cite{avila2017dronenavigator} employs drones for VIP navigation using a digital (Bluetooth) and a physical tether, while Zayer et al.~\cite{al2016exploring} utilize rotor sounds to guide VIPs in a controlled setting, with a human manually flying the drone. 
Our proposed solution uses drones that can autonomously follow the VIP wearing a \textit{hazard vest} and provide outdoor situation awareness~\cite{suman2023chi}, e.g., guiding them to walk along a collision-free path, notifying them of obstacles on the pavement, approaching intersections, etc. (Fig.~\ref{fig:teaser1}). These can complement accessibility technology based on smartphones and wearables. A key requirement for autonomous navigation of the drone and the VIP is to accurately \textit{estimate the distance} from the drone to the VIP, and to other potential obstacles, using just a monocular camera feed present in even low-end quadcopters and no other sensory input. Robustly achieving this is the focus of this article. These depth estimates can then be used by obstacle avoidance and path planning algorithms~\cite{10802577} to control the drone and offer cues to the VIP, as future work.

%% #########################################################
\paragraph{Challenges and Gaps}
Assistive solutions for VIPs should be affordable, portable and easy to use. Drones with stereo cameras are costlier, available only in professional drones, and require manual calibration for accuracy~\cite{scharstein2002taxonomy}. RGB-D depth sensors that rely on structured light or time-of-flight suffer in bright sunlight, have limited range, and tend to be bulkier and power-hungry~\cite{zhang2012microsoft}. Instead, monocular cameras are widely available in consumer drones, are portable weighing just 100g, and our proposed methods help ``auto-calibrate'' them for distance estimation, simplifying setup.

Monocular cameras lack true depth perception, and require post processing to estimate the absolute distance from the drone (camera) to the objects in the scene. There is an inherent scale ambiguity where the actual size and distance to objects cannot be distinguished without assumptions about object size or additional sensors~\cite{mur2017orb}.
State-of-the-art (SOTA) methods, such as \midas~\cite{ranftl2020towards} and Monodepth2~\cite{monodepth2}, generate a \textit{monocular depth map} (Fig.~\ref{fig:teaser2}) using DNN models from a video frame. But they only provide \textit{relative distances} for the pixels of the objects in the image rather than the \textit{absolute distances}. 
ZoeDepth~\cite{zoedepth} obtains scale and shift parameters to retrieve the true distances from the pixel depth scores, but is biased towards its training data and not generalizable.
Some methods~\cite{ranftl2020towards} exhibit higher errors at longer ranges, or under poor/harsh lighting~\cite{ranftl2021vision}.

Others~\cite{al2017obstacle} propose an obstacle detection algorithm by analyzing the change in the pixel-area of the approaching obstacles. Lee, et al.~\cite{lee2021deep} use DNNs over monocular feeds to navigate UAVs among trees in a plantation. They classify obstacles as critical or low-priority based on the height of the bounding boxes of the detected obstacles (trees). But these geometric methods cannot be generalized to estimate the absolute distances to the objects, required for VIP navigation. Lastly, given the constrained edge-computing on-board or co-located with drones, methods~\cite{wang2024duel,10610042} that require complex-calculations for each pixel in the frame can be intractable for real-time decisions.

%% #########################################################
\paragraph{Proposed Approach}
To address these gaps, we present \textit{NeoARCADE}, a \uline{No}v\uline{e}l \uline{A}pproach for \uline{R}obust \uline{CA}libration for \uline{D}istance 
\uline{E}stimation, which we refer to in short as ``\neo'' in the rest of the article. 
\textbf{\neo} is a generalizable method to estimate distances for any object in a monocular video frame, leveraging the depth map and object bounding-boxes generated by existing DNN models. 
We propose strategies for score normalization and depth parameter estimation that translate the relative distances in depth maps into an accurate absolute distance estimation to diverse objects proximate to VIP, using limited ground truth knowledge.
Hyperparameters of \neo trained on one VIP generalize well to others, and to other objects in the scene.
We perform recalibration to adapt to different lighting and adversarial conditions, such as VIP image being cropped.
We also use pixel sampling techniques within object bounding boxes to avoid costly computing on all pixels, and ensuring robustness to non-standard object orientations and geometries.

We also propose two alternative data-driven baselines for distance estimation. First, is a \textbf{regression}-based supervised model that uses outputs from an object detection model and fits a linear equation to estimate the distance based on prior training data. This is limited to just VIP distance estimation but is accurate, and also used by \neo to recalibrate under dynamic scenarios. Second, is a \textbf{geometric} model that estimates the distance to the VIP and to other objects using a pinhole camera model, homogeneous transformations, and the outputs from the object detection model. These are visualized in Fig.~\ref{fig:teaser}, where \textit{Regression} accurately estimates a distance of $2.99m$ to the VIP standing at a ground truth distance of $3m$ but works only for VIPs. \neo estimates a shorter distance to the VIP at $2.43m$, but performs better than Geometric for other objects, with more accurate distances of $4.01m$ to the bystander and $3.8m$ to the car compared to higher values by Geometric, while the respective ground truth distances are $4m$ and $4.6m$, respectively.
  
%% #########################################################
\paragraph{Representative Results against SOTA}\label{sec:comparison-with-sota}

\begin{figure} 
    \centering
    \begin{minipage}{0.47\columnwidth}
        \includegraphics[width=1\columnwidth]{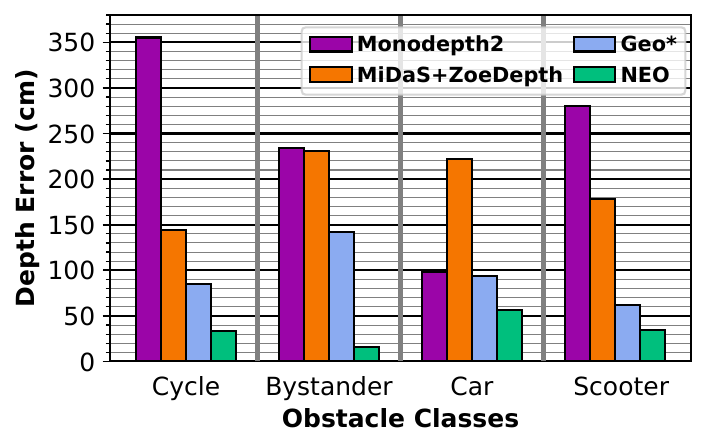}
    \caption{Comparison of \neo and Geo* with SOTA depth map approaches for distance estimation.}% using monocular drone camera
    \label{fig:baselines-comparison} 
    \end{minipage}
    \quad
    \begin{minipage}{0.47\columnwidth}
        \includegraphics[width=1\columnwidth]{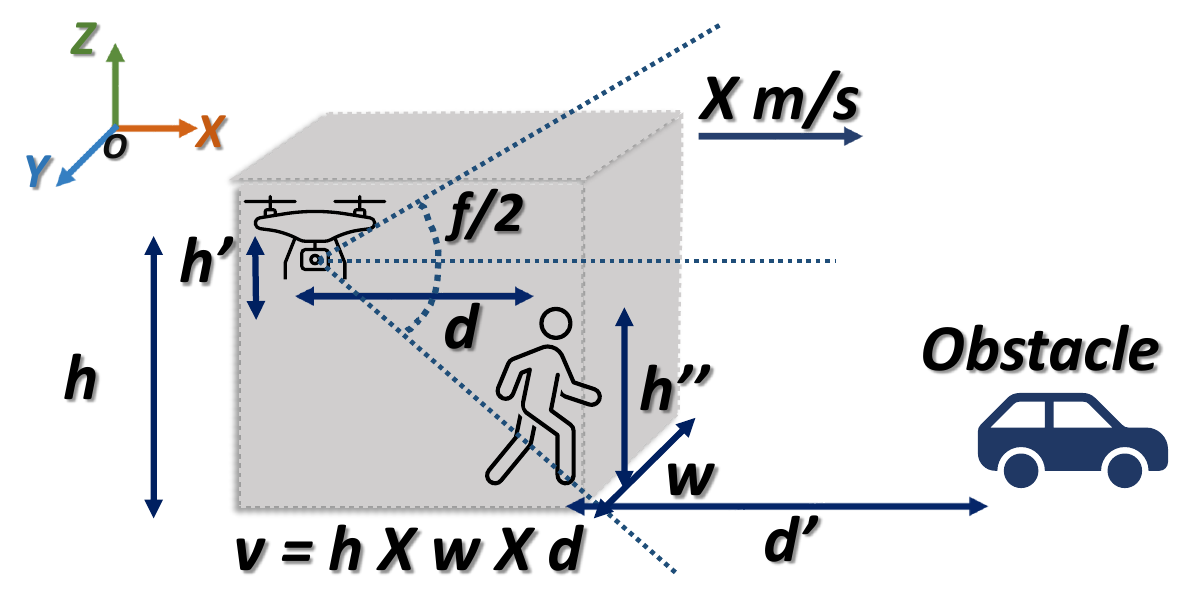}
    \caption{Illustration of different distance components involved in our distance estimation model.}
    \label{fig:illustration}
    \end{minipage}
\end{figure} 

We briefly report the performance of \neo relative to two other SOTA depth map-based approaches, \textit{MonoDepth2}~\cite{monodepth2} and \textit{\midas+ZoeDepth}~\cite{ranftl2020towards,zoedepth}, 
and the \textit{Geometric*} baseline (Geo*), over a larger set of drone video feeds. 
Monodepth2 provides a script to estimate the real depth using the Eigen split KITTI dataset~\cite{kitti_dataset} for calibration, and generates a median distance-scaling value of $33.26$, which we use to obtain the absolute distance. \textit{ZoeDepth} introduces metric bin modules for specific domains (e.g., indoor or outdoor) that compute per-pixel depth bin centers and are fine-tuned to predict metric depth using real-world measurements. \midas's decoder is passed to the metric bins module of ZoeDepth to get the absolute distance. As we discuss later, \neo adapts Monodepth2 using robust depth averaging and dynamic calibration.

Fig.~\ref{fig:baselines-comparison} compares the median \textit{absolute depth error (in cm)} for $3000$~video frames, from drone videos collected in our university campus as shared in Table.\ref{table:video-dataset-summary}. We report results for four common classes from the YOLOv8 model we use for object detection -- bystander (person who is not VIP), bicycle, motor-scooter and car. The objects are $2$--$4m$ from the drone. \neo has the least error of $60cm$ for cars and goes as low as $16cm$ for a bystander. This is up to $5.3 \times$  better than Geo*, $14.3 \times$ better than \midas and up to $14.6 \times$ than Monodepth2. The high depth errors, sometimes running into meters, make these alternative distance estimation methods impractical for guiding VIPs to avoid urban obstacles.

%% #########################################################
\paragraph{Contributions} 
We make the following key contributions in this article: 
\begin{enumerate}[leftmargin=*]
 \item We \textit{formally define the problem} of estimating the absolute depth from the drone camera to objects in the scene using monocular depth maps to support the needs of VIP navigation (\S~\ref{sec:problem-statement}).
\item We introduce two light-weight baseline methods to solve this: supervised \textit{regression} learning, and \textit{geometric} using a pinhole camera model, using limited ground-truth information (\S~\ref{sec:baselines}).
 \item We propose \neo, a generalizable and compute-efficient method to estimate distances to any object in an image frame using the monocular depth map for the scene (\S~\ref{sec:distance-estimation-neo}). We further develop robust calibration techniques to handle dynamic and adversarial conditions.
 \item We collect diverse ground truth dataset in our  campus that captures different object classes for static and dynamic scenarios (\S~\ref{sec:eval-dataset-collection}). We use these 
 for a rigorous evaluation of the different proposed methods, and compare them against Monodepth2 and MiDaS+ZoeDepth SOTA approaches, to highlight their relative accuracies, robustness, and limitations (\S~\ref{sec:evaluation}). 
 \end{enumerate}

\noindent We also compare our work with literature in \S~\ref{sec:related-work} and offer our conclusions and future work in \S~\ref{sec:conclusions}.

To the best of our knowledge, \neo is the first work that offers a lightweight and generalized approach to estimate absolute distances to objects in the image frame captured by a monocular camera using a data-driven approach. This article builds upon our prior work, \textit{Ocularone}~\cite{suman2023chi}, that motivates the use of UAVs to guide VIPs in urban environments, and a preliminary approach to obstacle avoidance~\cite{sumaniros}, that identified regions in the frame with minimal pixel density for detecting objects.
However, neither address distance estimations to objects, which is the focus of this article.

%% #############################################################################################
\section{Problem Statement} \label{sec:problem-statement}

We first describe the geometric model for the drone and VIP environment to help define the different distance-measures used in our navigation model. We consider a 3D volume with dimensions $h,d,w$, where $h$ is the height of the drone from the ground, $d$ is the longitudinal distance of the drone from the VIP, and $w$ is the width of lateral free space that the VIP requires for unobstructed motion (Fig.~\ref{fig:illustration}). $h'$ is the vertical offset between the top of the VIP with height $h''$ and the current height of the drone. For simplicity, we assume that the orientation of the drone with respect to the VIP is fixed, and the heading of the VIP coincides with the heading of the drone. 

\paragraph{Relative Positioning of the Drone}

The height offset ($h'$) and the distance ($d$) of the drone relative to the VIP is chosen such that the VIP is always in the Field of View (FoV) of the drone's onboard camera. If $f$ is the FoV of the camera, we can ensure that at least the top of the VIP's head is visible to the drone by setting the offset and height as:
\setlength{\belowdisplayskip}{0pt} \setlength{\belowdisplayshortskip}{0pt}
\setlength{\abovedisplayskip}{0pt} \setlength{\abovedisplayshortskip}{0pt}
\begin{equation}\label{eqn:tan}
    \tan{\Big(\frac{f}{2}\Big)} = \frac{h'}{d}
\end{equation}
  
This however is not adequate for accurate tracking of the VIP for two reasons: (1) There is a variability of speed of the VIP during the movement, which requires a range to be set, and (2) the drone needs to capture a significant portion of the VIP height to be able to follow them with these variations. Instead, we define a range of valid values for $h' \in [h'',h_{max}]$ and $d \in [d_{min},d_{max}]$, such that we have a straight line with coordinates $(h'',d_{max}),(h_{max},d_{min})$ satisfying Equation~\ref{eqn:tan}. We set $h_{max}$ and $d_{min}$ as the point on the straight line where roughly \textit{two-thirds} of the person starting from head is visible to the drone. We consider $d_{max}$ as the farthest distance the drone can go behind the VIP such that it can detect obstacles up to a minimum distance $d'$ ahead of the VIP. For practical purposes, we bound the values of $d_{min} \geq 2~m$ and $d_{max} \leq 4~m$ to stay within safe proximity limits.

\paragraph{Visibility Ahead of the VIP to Detect Obstacles}
 
Since our work focuses on human-in-the-loop paradigm, it becomes crucial to estimate distances to all the objects in the frame which is within a tunable \textit{safety-margin distance} ($d'$) on the path directly ahead of the VIP that needs to be obstacle-free while they navigate. This is necessary for a smooth navigation experience for the VIP~\cite{cloutier2022topical}.

This depends on multiple factors such as: (i) \textit{detection time} of the obstacle, between the camera observing it in its video frame to the processing of the frame using DNN models to identify the obstacle, (ii) \textit{human reaction time}, which is the average time required for the VIP to comprehend and respond to an audio/sensory cue from our system when notifying them of an obstacle, and (iii) speed at which the VIP is walking ($x~m/s$). Again here, based on our practical experiments, we set two distance-ranges for $d'$. Objects $\leq 1.5m$ from the VIP, which can pose an imminent danger to the VIP, and objects between $1.5$--$5.5m$ from the VIP are with a lower risk.

Using the above setup, we define our problem statement as: 
``\textit{Accurately estimate the distance from the drone to the VIP and to other objects present in the monocular video feed from the front-facing drone-camera, when the drone is following the VIP from a distance $d$ and at a height $h$ from the ground, with priority given to nearby-objects.}''.

%% #############################################################################################
%% #############################################################################################
\section{Distance Estimation using Baselines} \label{sec:baselines}
The input to all our models is the video frame image  from the drone camera annotated with the list of bounding boxes and classes for objects (including the VIP) present in the frame, generated by an object detection DNN model such as YOLO~\cite{yolov8_ultralytics}.
We propose two CV-based baselines to estimate the distance to a VIP ($d_{V}$) given a \textit{video frame} of the scene and the \textit{bounding box of the detection of the VIP} within that image (Fig.~\ref{fig:teaser1}). This is then extended to distance estimation to other objects.

%====================================================
\subsection{Regression-based Model for VIP}\label{sec:supervised-learning}
Here, we adopt a simple supervised learning method, \textit{Regression}, that uses ground-truth distances ($d^i$) collected from the drone to the VIP for several video frames ($i$) at different static distances ($d_{V}$). We train a linear regression model on the input features of the VIP detection bounding box: \textit{width} ($w_{b}^i$), \textit{height} ($h_{b}^i$), and \textit{area} ($A_b = w_{b}^i \times h_{b}^i$), detected for the VIP in each frame $i$.
\begin{equation}\label{eqn:regression}
    d^i = a \cdot w_{b}^i + b \cdot h_{b}^i + c \cdot A_{b}^i \qquad \forall d^i \in d_V
\end{equation}

\begin{figure}[t]
    \centering
    \begin{minipage}{0.50\columnwidth}    
    \centering
    \footnotesize
    \includegraphics[width=1.0\columnwidth]{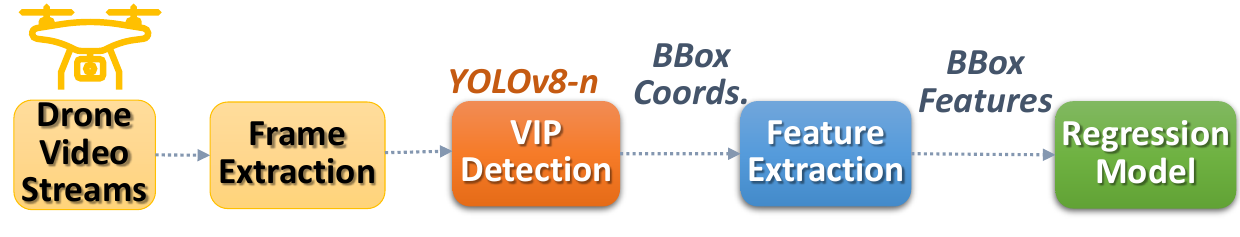}
    \caption{Workflow for distance estimation using Regression-based Model.}
    \label{fig:regression-workflow}
    \setlength{\tabcolsep}{2pt}
    \caption{Train/Test Dataset for Regression Model.}
    \label{tab:regression_train_data}
    \begin{tabular}{c || r | r}
    \hline
        \bf $d_{V}$~(in m) & \bf \makecell{\# Training\\ Images $(i)$} & \bf \makecell{\# Testing\\ Images} \\
        \hline        
        \hline
        $2$ &  $893$  & $597$\\ 
        \hline
        $2.5$ & --- & $590$\\
        \hline
        $3$ &  $921$ &  $589$\\
        \hline
        $4$ & $963$ & $594$\\
        \hline
    \end{tabular}
        \end{minipage}
    \quad
    \begin{minipage}{0.46\columnwidth}
        \centering
    \includegraphics[width=1\columnwidth]{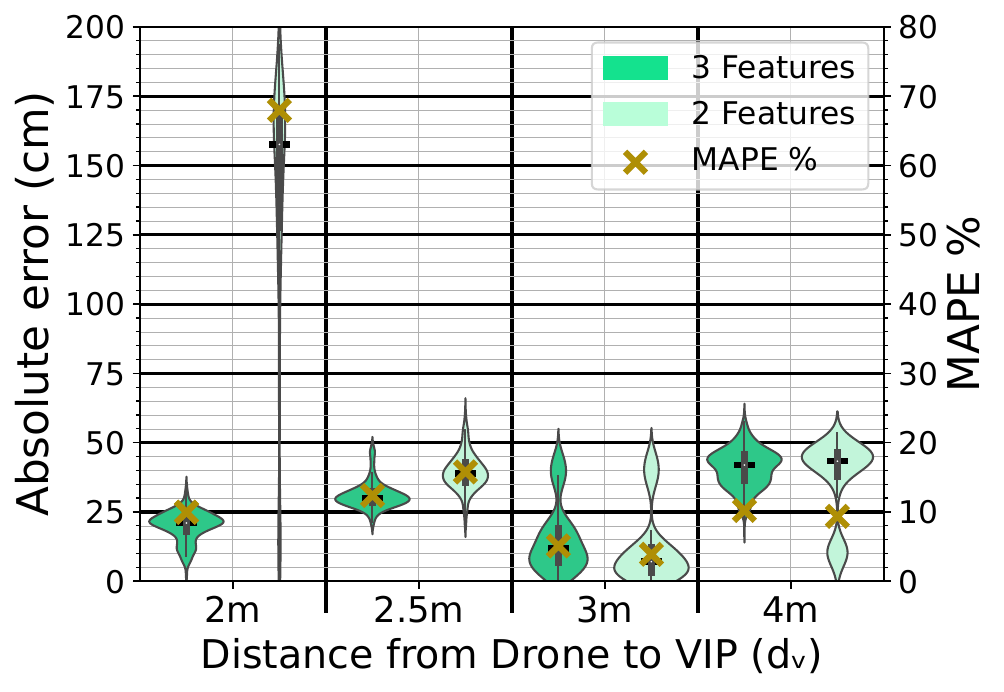}
    \caption{Performance of Linear Regression models for VIP distance estimation.}
    \label{fig:regression-benchmark}
    \end{minipage}
\end{figure}

We solve this over different frames $i$ collected at different distances to fit the coefficients $a,b,c$, which are calibrated for the specific VIP and the drone camera (Fig.~\ref{fig:regression-workflow}). Later, given the bounding box of the VIP in a frame at an unknown distance $d$, we can solve this equation to obtain $d$.

We evaluate two variants: one trains the linear regression model using all $3$ features while another uses just $w_b$ and $h_b$ as features. We validate both models on the training and testing dataset in Table~\ref{tab:regression_train_data}, whose $d_V$ distances fall in the range of expected distances between the drone and the VIP. Fig.~\ref{fig:regression-benchmark} compares the accuracy of the predictions using the \textit{Absolute Error (in cm)} for each distance range on the left-Y axis, along with \textit{Mean Absolute Percentage Error (MAPE\%)} shown on the right-Y axis. The model with 3 features has a tighter distribution and an overall median error of $27.2~cm$ compared to the one with just 2 features, which has a median error of $40.9~cm$. The former is much better for $d_V < 3~m$ and comparable for $d_V \geq 3~m$. So, we use the 3-feature model in later sections.

While the regression model works well for VIPs, it is not generalizable for distance estimation to other objects since it is impractical to obtain the ground-truth bounding box features at different known distances for a large class of objects, required to train this model.

%%===================================================
\subsection{Geometric Model}\label{sec:geometrical-model}

\begin{figure}[t]
    \centering
    \includegraphics[width=0.75\columnwidth]{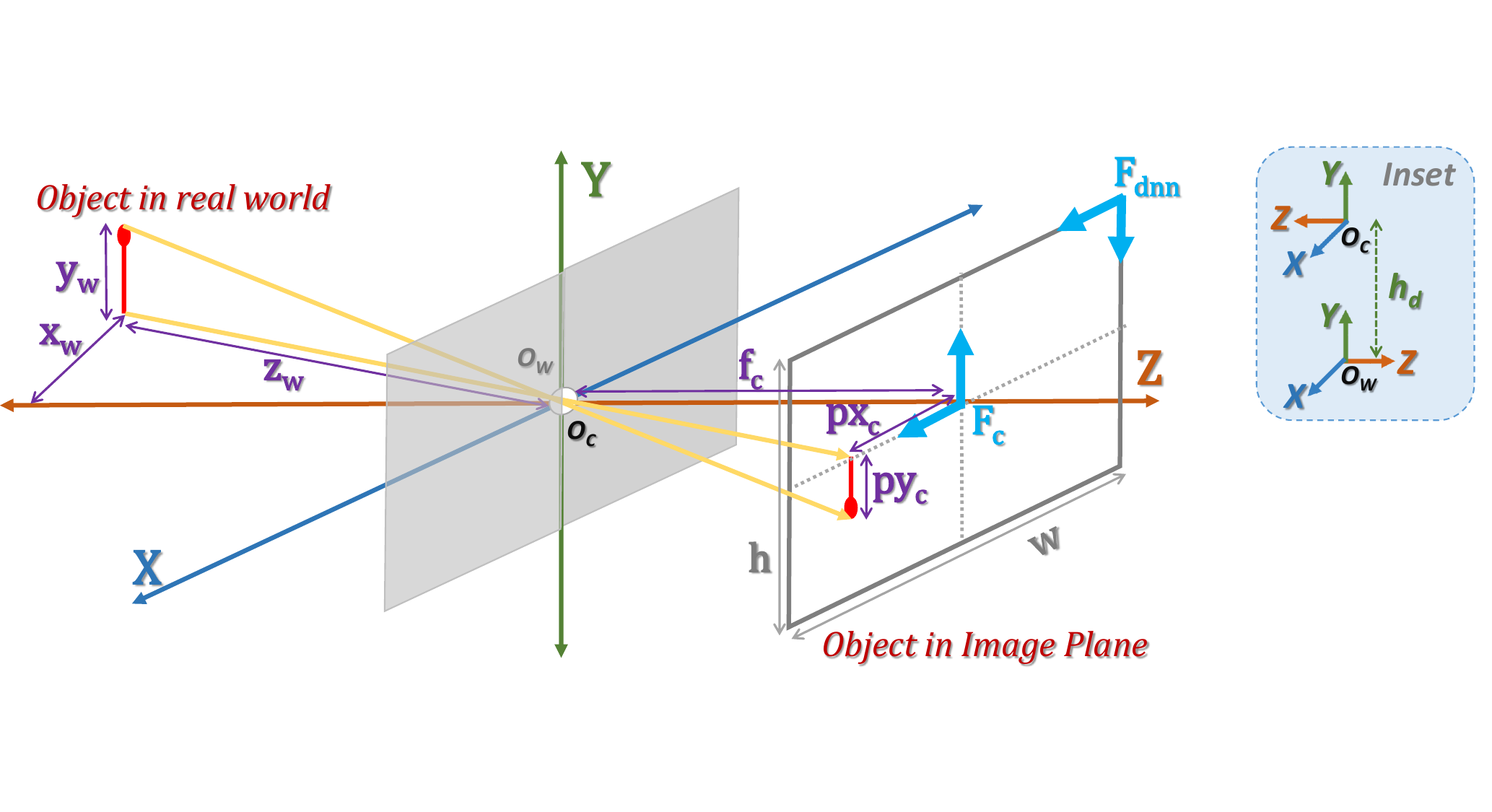}
    \caption{Geometric model of pinhole camera for a real-world object projected to image plane. }
    \label{fig:3d-pinhole}
\end{figure}

A pinhole camera model (Fig.~\ref{fig:3d-pinhole}) captures the geometry of an object in the real world projected by the camera lens onto the image plane~\cite{Sturm2014}. Using basic geometric transformations, we can determine the distance between the VIP (hazard vest) and the drone (camera lens), given the pixel height of the vest, the actual height of the vest, and the focal length of the lens, while accounting for the coordinate systems of the real world, image plane and object detection DNN model. 

Let $\mathcal{O}_w$ be the origin of the world coordinate frame ($\mathds{W}$) and $\mathcal{O}_c$ be the origin of the camera coordinate frame ($\mathds{C}$). In Fig.~\ref{fig:3d-pinhole}, we assume that $\mathcal{O}_w$ coincides with $\mathcal{O}_c$, and is at the pinhole (camera lens, shown by a white circle in center plane). Then, the object in the real world with a height of $y_w$ meters (red upright line on left) is projected and inverted by the camera lens onto the image plane of the camera with a pixel height of $py_c$ (red inverted line on right). The horizontal offset along the X axis and the distance from the lens are $x_w$ and $z_w$ for the real-world object and $px_c$ and $f_c$ for its projected image, where $f_c$ is the focal length of the camera. $z_w$ is the \textit{distance to be estimated} from the drone to the vest.

The image plane is centered at $\mathds{F}_c$ and the image has a height $h$ and width $w$ in pixels. However, the object's bounding box returned by the object detection model uses the top-left corner of the frame as its origin, $\mathds{F}_{dnn}$. Hence, the pixel coordinate $({px}_{bb},{py}_{bb})$ of the object bounding box ($bb$) relative to $\mathds{F}_{dnn}$ is converted to a pixel coordinate $({px}_c,{py}_c)$ relative to camera $\mathds{F}_c$: 
\begin{equation}\label{eqn:origin-transformation}
    {px}_c = \big({px}_{bb} - \frac{w}{2}\big)
    \qquad\qquad
    {py}_c = -\big({py}_{bb} - \frac{h}{2}\big)
\end{equation}

Next, we generalize this to the camera being at an elevation from the world coordinate since the drone is flying. We adopt the common convention~\cite{Hartley2004} where the Z-axis of the camera coordinate frame is aligned on the negative Z-axis of the world coordinate. If $h_d$ is the height of the drone (Fig.~\ref{fig:3d-pinhole} Inset), $\mathcal{O}_c$ translates $h_d$ in the positive Y-direction with respect to $\mathcal{O}_w$ using the homogeneous coordinate transformation matrix:
\begin{equation}\label{eqn:homogeneous}
\textbf{g} = \begin{bmatrix}
1 & 0 & 0 & 0\\
0 & 1 & 0 & h_d\\
0 & 0 & -1 & 0 \\
0 & 0 & 0 & 1
\end{bmatrix}
\end{equation} 

Let $\mathds{P}_w = (x_w,y_w,z_w,1)$ be the homogeneous coordinates in $\mathds{W}$. Then, we calculate $\mathds{P}_c = (x_c,y_c,z_c,1)$ as: 
\begin{equation}\label{eqn:homogeneous-short}
    \mathds{P}_c = \textbf{g}\mathds{P}_w
\end{equation}
where $\mathds{P}_c$ denotes the homogeneous coordinates in $\mathds{C}$, where $1$ is the scaling factor. Using the pinhole camera model~\cite{Sturm2014} and a camera focal length of $f_c$, the 2D coordinates of the projection of $\mathds{P}_c$ on the image frame $\mathds{F}_c$ are:
\begin{equation}\label{eqn:pinhole}
\begin{bmatrix}
    px_c \\ py_c
\end{bmatrix} = \begin{bmatrix}
    f_c\times(\frac{x_c}{z_c}) \\
    f_c\times(\frac{y_c}{z_c})
\end{bmatrix}
\end{equation}
Substituting $x_c$ and $y_c$ in Eqn.~\ref{eqn:pinhole} using Eqn.~\ref{eqn:homogeneous-short} and assuming the focal plane is in front of the camera:
\begin{equation}\label{eqn:pinhole2}
    {px}_c = f_c\times\frac{x_w}{z_w}
    \qquad\qquad
    {py}_c = f_c\times\frac{(y_w + h_d)}{z_w}
\end{equation}

Next, we relax this to a scenario where the object is not a point object but has a height $h_o$ in the world coordinate $\mathds{W}$, with $h_o = (y_{w1} - y_{w2})$. Here, $y_{w1}$ and $y_{w2}$ are the top and bottom Y-coordinates of the object in the real world. Similarly, the height of the image of that object is $h_i = ({py}_{bb1} - {py}_{bb2})$, where ${py}_{bb1}$ and ${py}_{bb2}$ refer to the top and bottom pixel-coordinates of the image. On projecting $h_o$ in $\mathds{W}$ as $h_i$ on the camera focal plane in $\mathds{C}$, we get the equation to calculate $z_w$, the real-world distance from the object (e.g., VIP's vest) to the camera (drone) as:
\begin{equation}\label{eqn:final}
    z_w = f_c\times\frac{h_o}{h_i}
\end{equation}

\begin{figure}
    \centering
    \subfloat[Geometric Model]{\includegraphics[width=0.47\columnwidth]{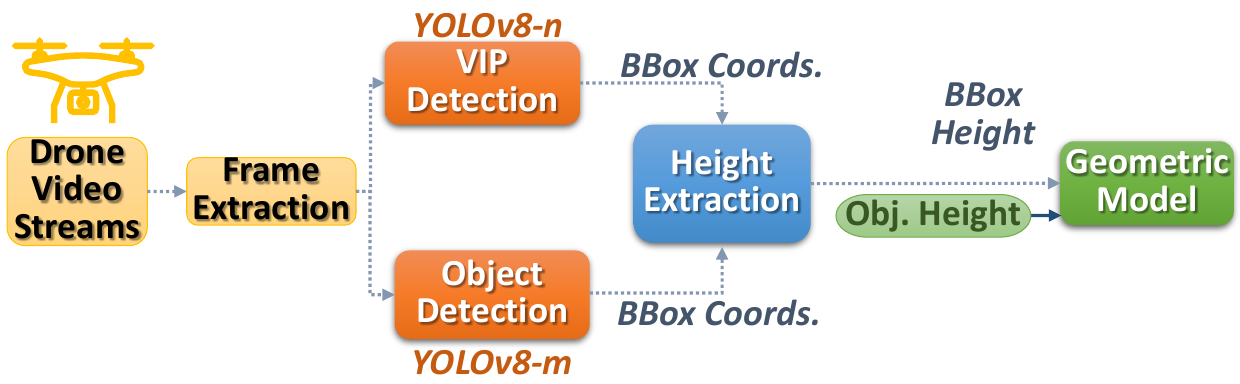}\label{fig:workflow-geometric}}\quad
    \subfloat[\neo]{\includegraphics[width=0.47\columnwidth]{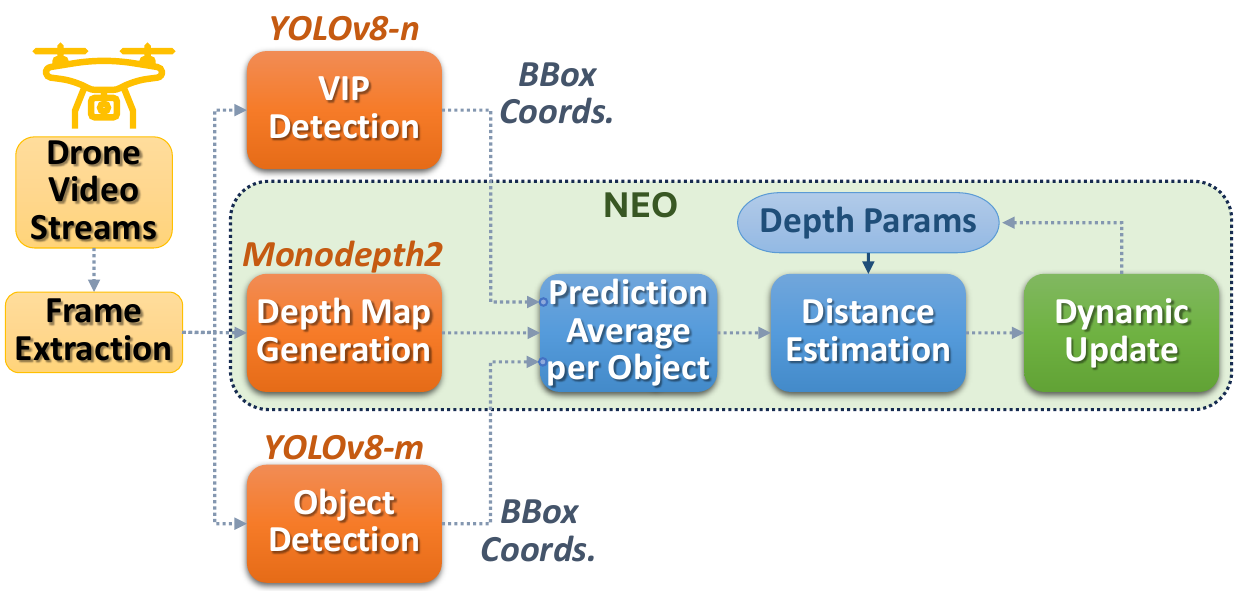}\label{fig:workflow-neo}}
    \caption{Workflow for distance estimation using Geometric and \neo methods.}
\end{figure}

When calibrating the geometric model (Sec.~\ref{sec:eval:calib}), we provide the height of the object, their bounding boxes, and the focal length of the camera -- found experimentally or provided from the drone datasheet. Then, for a given object in the scene with a known real-world height $h_o$ and a bounding box height $h_b$, we fit them in Eqn.~\ref{eqn:final} to calculate their distance from the drone, $d$ (Fig.~\ref{fig:workflow-geometric}).

We have two variants of this method based on the value of $h_o$ provided: (a) \textit{Geometric} that uses \textit{average height estimates} for an object class from a lookup table, and (b) \textit{Geometric*} that uses \textit{actual height} of the object that is in the frame. The former is easier to get for a larger class of objects while the latter is specific to heights of individual instances of the object, which is harder to secure.

We use the DJI Tello nano-drone in our experiments. Its datasheet does not provide focal length $f$ but only the FOV, $82.6^\circ$. Finding the focal length from the FOV is cumbersome. So we experimentally calculate $f_c$ using Eqn.~\ref{eqn:final} from several reference images collected by the drone. Specifically, we extract $\approx1000$ images from a short video where Tello flies at a constant height of $1.5~m$ and at $3~m$ from the VIP. The vest's height is given as $0.63~m$. This gives a distribution of the focal lengths over $1000$ images, which we find is bounded between quartiles, $Q1=1538$ and $Q3=1610$~pixels. We use the median focal length, $Q2=1592$~pixels, in subsequent sections. We confirm that this estimate works equally well for images at different drone heights and distances from the VIP.

%% #############################################################################################
\section{Distance Estimation using NeoARCADE}\label{sec:distance-estimation-neo}
Here, we propose a novel distance estimation approach, \neo, leveraging recent works on \textit{depth maps} using DNN for monocular images~\cite{monodepth2,eigen2014depth,ranftl2020towards}. Depth map DNN models score each pixel in a frame such that pixels estimated to be at the same distance from the camera have the same score. There exists some transformation function from the score to the real-world distance to the object represented by that pixel, but this needs to be discovered through \textit{calibration}. \neo performs this calibration by using: (i) a sample video frame with the VIP, (ii) the distance to the VIP in that frame provided as ground truth, and (iii) the depth map for that frame from a DNN model. The ground truth distance for the sample video frame can be measured once manually by the VIP (e.g., stand at $2m$ distance and take a picture), calculated using the geometric and/or regression model baselines for images with the VIP, or estimated using a colocated LiDAR. This calibration serves as the transformation function for the depth map, and converts the pixel scores to the real-world distances. Importantly, once calibrated, we can also use this function to find the distance to any other object in depth maps from the same DNN model generated for other frames of the same camera. Figure~\ref{fig:workflow-neo} illustrates the building blocks of \neo, which we describe next.

\subsection{Choice of Depth Map DNN Model}\label{sec:depth-map-models-choice}
We first need to choose our base DNN model that generates depth maps using monocular images. We select Monodepth2~\cite{monodepth2}, that uses a fully convolutional U-Net to predict the depth and per-pixel minimum reprojection loss, auto-masking stationary pixels, and multi-scale estimation to improve self-supervised depth estimation for stereo as well as monocular images. We have also evaluated alternate models: Adabins~\cite{bhat2021adabins}, U-Depth~\cite{yu2023udepth}, Marigold~\cite{ke2023repurposing}, MiDaS~\cite{ranftl2020towards} and DenseDepth~\cite{Alhashim2018}. However, only Monodepth2 and MiDaS exhibit a monotonic trend in depth scores when the distance of objects from the camera changes. Among these two, Monodepth2 is more reliable. 

Monodepth2 generates a $score$ for each pixel, indicating its relative inverse depth in the frame. The absolute real-world depth $d$ for a pixel given its $score$ is:
\begin{equation}\label{eqn:depth-map}
    {d} = m \cdot score + s
\end{equation}
where $m$ and $s$ are the \textit{scale} and \textit{shift} depth parameters. We extract frames from the drone video, downscale them from a resolution of $1280 \times 720$~p to $1024 \times 320$~p required by Monodepth2, and get the depth map for the scaled video frame (Fig.~\ref{fig:workflow-neo}). We correspondingly scale and overlay the \textit{bounding boxes} for the objects in the frame from YOLOv8, and extract the depth map images for each object within these bounding boxes. This gives us depth values for each pixel within the bounding box for the object. We then  perform depth score normalization, as discussed next.

\subsection{Depth Score Normalization per Object}\label{sec:score-normalization}

\begin{figure}
    \centering
    \includegraphics[width=0.85\columnwidth]{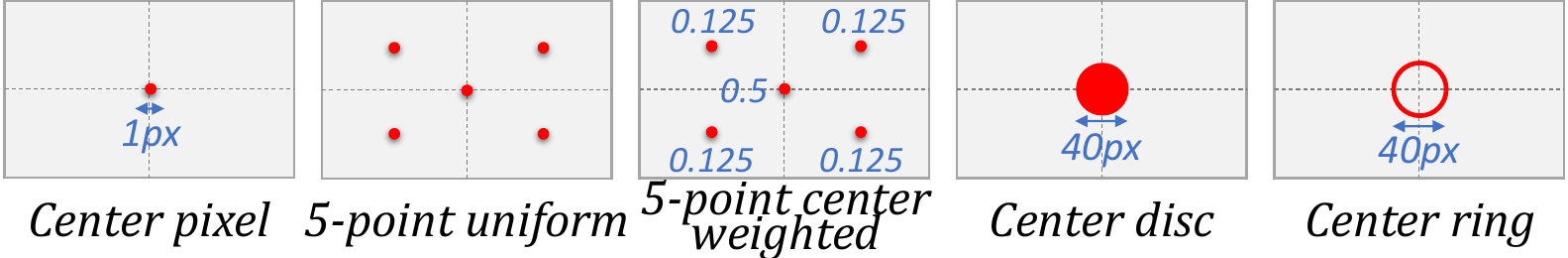}
    \caption{Depth score normalization techniques for an object's depth map.}
    \label{fig:neo-avg}
\end{figure}

The depth map scores for the pixels within an object can vary significantly within the bounding box, e.g., due to the object having a complex shape, non-uniform depth (car in Fig.~\ref{fig:teaser2}), variable lighting, etc. We propose several \textit{normalization techniques} (Fig.~\ref{fig:neo-avg}) to determine a single representative depth score for the object that is used in the calibration model to ascertain its actual depth.

The simplest approach is to select the \textit{center pixel} of the bounding box of the object as its representative pixel. But this can skew the results for complex objects if this pixel happens to fall on some background behind the object. We mitigate this by selecting 5-points, one at the center and one each at the center of each quadrant of the box. We the weight them equally (\textit{5-point uniform}) or giving higher weight to the center pixel (\textit{5-point center-weighted}). These avoids single-pixel errors.
Alternatively, we can also take the average of the depth scores of pixels within a \textit{disc at the center}, or average the circumference of this disc to give \textit{center ring}, e.g., of diameter 40px, to give more weight to the center. But, the center of the box is may not be the \textit{nearest portion} of that object from the camera -- important when we need to determine obstacles closer to the VIP. To address this, we can average the lowest $10$ percentile of the pixel depth scores in the box (nearest region), called \textit{Low Threshold (LT)}.
Lastly, we can compute the \textit{median} or \textit{mean} of the scores for all pixels in the box.

We conduct a study across various object classes positioned at different distances from the camera to evaluate these normalization techniques.
LT performs the best for our VIP use case consistently for different classes, and objects of different shapes and orientations.
E.g., for a bicycle placed $3m$ from the camera, the distances reported are $3.23m,\ 5.19m,\ 6.37m,\ 3.85m,\ 4.62m$ and \underline{$3.16m$} using the center pixel, 5-point center-weighted, 5-point uniform, disc at the center, center ring and \underline{LT} methods, respectively, with LT having the least error. LT is also robust to complex shapes and orientations since it considers the pixels \textit{nearest} to the drone, which form the immediate obstacle to the VIP, e.g., a car's bonnet seen in the bottom part of a frame may be facing the drone and hence closer than the top part of the frame. Hence, we use LT for subsequent experiments. As we discuss later, we complement this with a sliding window averaging technique across frames in a video to smooth out the changes in the normalized score.

\subsection{Estimation of Depth Coefficients for Calibration}\label{sec:depth-params-estimate}

In theory, we need scores for two pixels and their real-world distances to fit the scale ($m$) and shift ($s$) calibration coefficients in Eqn.~\ref{eqn:depth-map}. These can be obtained by taking two frames with the VIP present at two different known distances, and using the scores for the pixels of their vest. However, we find this to be highly sensitive to the specific video scenes from which the frames are chosen, the lighting conditions and the distances to the VIP.

We evaluate this calibration using videos for the VIP at $6$ different distances from the drone, and select different distance pairs to perform the fit:
$D=\{(2m,3m), (2m,3.5m), (2m,4m), (2.5m,3.5m),\allowbreak (2.5m,4m),\allowbreak (3m,4m)\}$. These pairs are chosen such that the difference in distances for each is $>=1m$, to cover sufficient spatial region.
We use multiple frames from videos at each distance-pair to fit Eqn.~\ref{eqn:depth-map} and return a distribution of $m$ and $s$ values for each; and pick their median. We perform this for different backgrounds and lighting conditions for the VIP, as described in \S~\ref{sec:eval:calib}, and choose the best combination of $m$ and $s$ from among them based on the lowest sum of absolute median errors and the lowest sum of positive and negative median errors.

\subsection{Dynamic Recalibration of Depth Coefficients} 
\label{sec:dynamic-update}

\begin{figure}
    \centering
    \includegraphics[width=0.7\columnwidth]{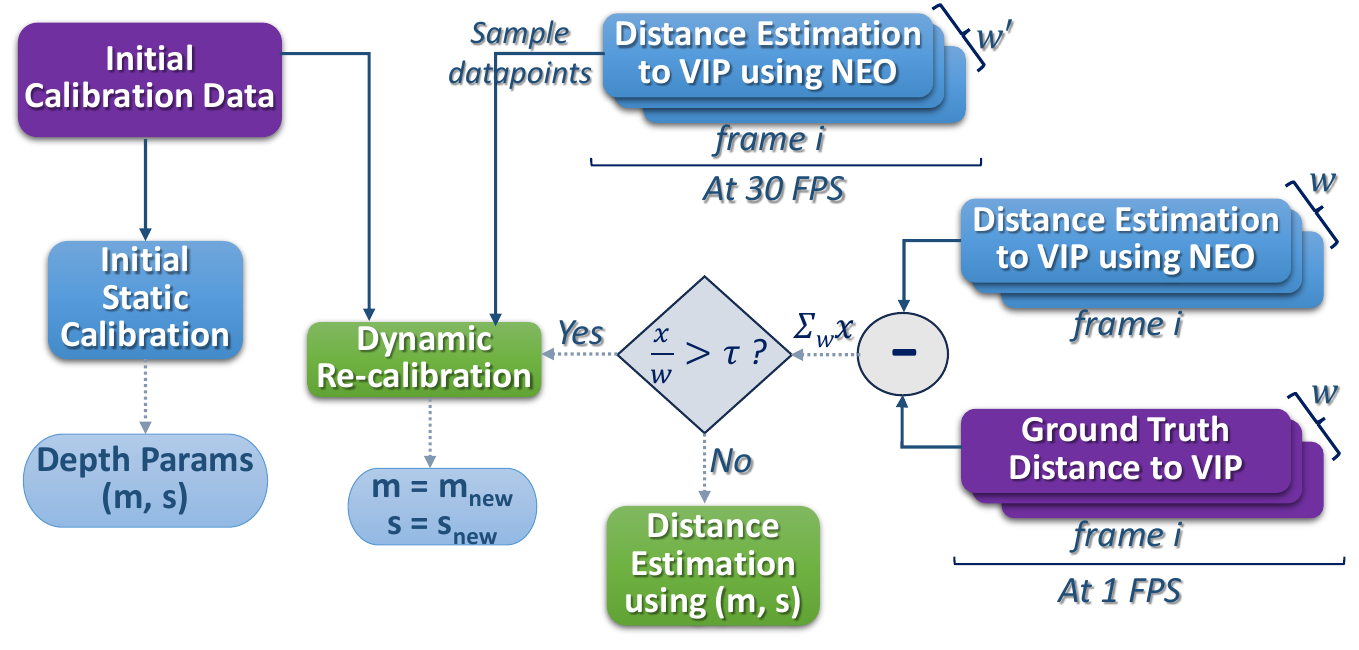}
    \caption{Dynamic Re-Calibration of Depth Parameters.}
    \label{fig:dynamic-recalibration}
\end{figure}

Depth maps can be affected by lighting and ambient conditions, as they affect the quality and accuracy of visual features captured by the camera. Changes in lighting can cause inconsistencies in feature extraction, while varying environmental conditions, e.g., sunlight, shadows or reflections, can distort depth estimates. These are especially critical in dynamic outdoor settings, where the scenes are constantly evolving. This can cause discrepancies between the expected and observed depth values. When this deviation rises beyond a threshold, it indicates that the calibration no longer works for the current ambient setting and a recalibration is required.

While the initial calibration is done offline, these ambient changes require recalibration to be done dynamically to evolve with changing scenes as the VIP  moves. We propose a novel mechanism relevant to the VIP scenario where it leverages the fact that the Regression method is accurate for VIP distance estimation under diverse conditions. This is used to dynamically detect distance errors, and recalibrate \neo's depth map coefficients, $m$ and $s$. For \textit{detecting} distance errors at runtime, \neo compares its distance estimates for the VIP against the Regression model's distance, which serves as ground truth, over a sliding window of frames. If this difference exceeds a threshold $\tau$ recalibration is triggered. 
During \textit{recalibration}, depth coefficients are updated by using both recent video frames from the current scene and initial frames used during offline calibration; the tunable parameter $\alpha$ decides the weightage for recent vs. initial frames. 
While the detection and recalibration are based on the VIP's image, the recalibrated parameters are used for all objects. This helps achieve robust and accurate distance estimates across scenes. Next, we discuss these in detail (Fig.~\ref{fig:dynamic-recalibration}).

\begin{algorithm}[t]
\caption{Dynamic Recalibration of Depth Coefficients}\label{algo:dynamic-recalibration}
\small
\begin{algorithmic}[1]
\Function{Recalibrate}{$w, w', \alpha, n_o, n_{new}, n_{orig}, \tau$}\\
    \Comment{\emph{$w$ and $w'$ are the window sizes of the sliding windows over the VIP images at $1$ FPS and $30$ FPS respectively. $n_o$ is the number of frames in $n_{orig}$ and $n_{new}$ is the number of data points sampled from the recent $w'$ $\times$ $30$ frames. $\alpha$ is the weight given to the original parameters}} 
    \State Initialize $R[~] = 0$, $D[~] = 0$ and $T[~] =0$ \Comment{\emph{Initialize regression buffer, depth buffer as empty list of size w and train buffer depth as empty lists of size w'}}
    \State Initialise flag = $0$, $i,j = 0$
    \For {each new image}
        \If {$j < 5$}
        \State $R[j]$.append($d_G$) at $1$ FPS 
        \State $D[j]$.append($d_{\neo}$) at $1$ FPS
        \State $T[(j\times30)+0 \dots (j\times30) + 29]$.append($d_G$) at $30$ FPS
        \Else     
        \If{flag == 0}
            \State $i = j \mathbin{\%} w$
            \State $\mathds{R}[i] = d_G$ at 1 FPS
            \State $\mathds{D}[i] = d_{\neo}$
            \State $T[(j\times30)+0 \dots (j\times30) + 29] = d_G$ at $30$ FPS
        \EndIf
        \EndIf

    \If{$\frac{\sum\limits_{i=(0,w)} \left|{d_{G}^{i} - d_{Neo}^i }\right| }{w} > \tau $}\Comment{\emph{Detection}}
    \State flag == $1$
    \EndIf
        
    \If{flag} \Comment{\emph{Re-computation}}
    \State Calculate $n_{new} = \frac{n_o}{\alpha} - n_o$
    \State Sample $n_{new}$ data points from T and append to $n_o$
    \State Perform linear fit over $n_{orig} + n_{new}$ samples
    \EndIf
    \State $j+ = 1$
    \EndFor
\EndFunction
\end{algorithmic}
\end{algorithm}

\subsubsection{Detection} 
We first decide if the existing coefficients need to be re-calculated for a given video scene.
We compare the accuracy of the distance estimation to the VIP given by \neo, relative to the ``ground truth'' distance to VIP, provided by \textit{Regression} or \textit{Geometric} models that are more reliable for the VIP, or any other sources. We maintain a sliding window over the VIP image frames at $1$~FPS, with a window duration of $w$ seconds, which stores the depth predictions to the VIP by \neo ($d_{Neo}$) and by a ground truth model ($d_{G}$). If the average absolute distance error over this window using \neo exceeds a certain threshold $\tau$, we trigger a recalibration of the depth coefficients: $\frac{\sum\limits_{i=(0,w)} \left|{d_{G}^{i} - d_{Neo}^i }\right| }{w} > \tau$.
This approach ensures that the recalibration is triggered only when necessary. Using a sliding window to track errors over time captures persistent inaccuracies rather than transient noise, and ensures a robust and adaptive performance in dynamic environments.

\subsubsection{Recalibration}
Once \neo detects the need to re-calibrate, it triggers the process to adjust the depth coefficients, as shown in Algorithm~\ref{algo:dynamic-recalibration}. Towards this, we additionally maintain a sliding window of the VIP frames at $30$~FPS with a window duration of $w'$ seconds, which stores the normalized depth scores and the ground truth distances to the VIP. Similar depth scores and their ground truth distances to the VIP are also maintained for the $n_{o}$ frames used for offline initializing of the coefficients, collected from the ``best'' distance-pairs (\S~\ref{sec:depth-params-estimate}). 

We then calculate $n_{new} = \frac{(1-\alpha)}{\alpha}\cdot n_o$, where $\alpha$ is a tunable parameter, giving a weightage of $\alpha$ to the frames used to generate the original parameters and a weightage of $1-\alpha$ to recent frames. We use this to sample $n_{new}$ data points from the recent $w' \times 30$ frames and append them to the original set of frames. We perform a linear fit over these ($n_{orig} + n_{new}$) samples to recompute $m$ and $s$.  This approach ensures a balance, allowing sufficient new data to influence the recalibration and account for recent environmental changes, while also preserving the generality offered by the original calibration data and preventing a skew towards only recent frames that may be a transient phase. We set the hyper-parameters $w$, $w'$, $\alpha$ and $\tau$ used for evaluation in \S~\ref{sec:eval:calib}.

%% #############################################################################################
%% #############################################################################################
\section{Experimental Setup and Datasets}\label{sec:setup}

\begin{table}[t]
\caption{DNN Models and Inference Performance on NVIDIA Jetson Orin Nano.}
\centering
\footnotesize
\renewcommand{\arraystretch}{0.9} 
\begin{tabular}{c | c | c }
\hline
\textbf{Application} & \textbf{DNN Model}  & \textbf{Infer. time/frame (ms)}\\ 
\hline
\hline
VIP Detection & RT YOLO v8 (nano) & 35\\ 
\hline
Object Detection & YOLO v8 (medium)  & 68\\  
\hline
Monocular Depth Map & Monodepth2 & 15\\ 
\hline
\end{tabular}
\label{table:dnn-config}
\end{table}

\subsection{Implementation and Setup}
The distance estimation approaches discussed above have been implemented in \textit{Python}. We use \textit{PyTorch v2.0.0}
for invoking the various DNN models (Table~\ref{table:dnn-config}) for inferencing over the video streams to generate outputs that feed into our distance estimation and calibration models.Videos used in these experiments for training and testing are recorded using a \textit{DJI Tello}~\cite{tello} nano quad-copter, which weighs $80~g$ with battery and has an onboard $720$p HD monocular camera that generates feeds at $30$ frames per second (FPS). The frames are extracted using \textit{moviepy} Python library.
We use an \textit{Nvidia Jetson Orin Nano} developer kit~\cite{jetsonorinnano} as our accelerated edge device. It has a GPU with $1024$ Ampere CUDA cores, $32$ tensor cores, a $6$-core Arm Cortex-A78AE CPU, and $8$~GB RAM shared by CPU and GPU. It is $176$~g in weight, $10 \times 7.9 \times 2.1~cm$ in size and can be powered by a portable USB powerbank. It is compact enough to be carried by the VIP in their backpack or purse. Its inferencing times per frame for the DNN models we use are shown in Table~\ref{table:dnn-config}. To detect the VIP in the frames, we retrain the YOLOv8-nano model, and achieve an accuracy of 99.4\% on normal images and 96.0\% on adversarial images such as low light, blurred images, etc.~\cite{raj2025ocularonebenchbenchmarkingdnnmodels}.

%==========================================
\subsection{Distance Evaluation Video Dataset}\label{sec:eval-dataset-collection}
For our evaluations, we collect videos of various real-life and managed scenes encountered by a student serving as a VIP-proxy walking on our campus. The videos are recorded using the DJI Tello drone at a height of $1.5$~m and at $30$~FPS. These scenes exhibit diversity along several dimensions for a realistic evaluation. To accurately record the ground-truth distance between the drone and the VIP-proxy while moving, we periodically use a measuring tape held by the VIP at one end and the drone operator at the other. We also use traffic cones as reference markers at known distances from objects in the scene to correlate with the object--VIP distances as the VIP passes them. These ensure consistent ground-truth distance measurements along with natural movement by the VIP.

\begin{figure}[t]
\centering%~
  \subfloat[\texttt{AM\_Simple}]{%
    \includegraphics[width=0.33\columnwidth]{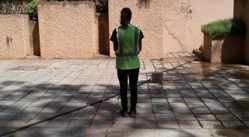}
   \label{fig:am-simple}
  }%
  \subfloat[\texttt{AM\_Complex}]{%
   \includegraphics[width=0.33\columnwidth]{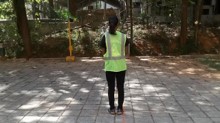}
    \label{fig:am-complex}
  }%
  \subfloat[\texttt{PM\_Complex}]{%
   \includegraphics[width=0.33\columnwidth]{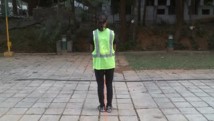}
    \label{fig:pm-complex}
  }%
\caption{A proxy for VIP standing at $4m$ from drone's camera depicting different scenes used for evaluation.}
\label{fig:scenes-description}
\end{figure}
\begin{figure}[t]
\centering%~
  \subfloat[Bicycle]{%
   \includegraphics[width=0.24\columnwidth]{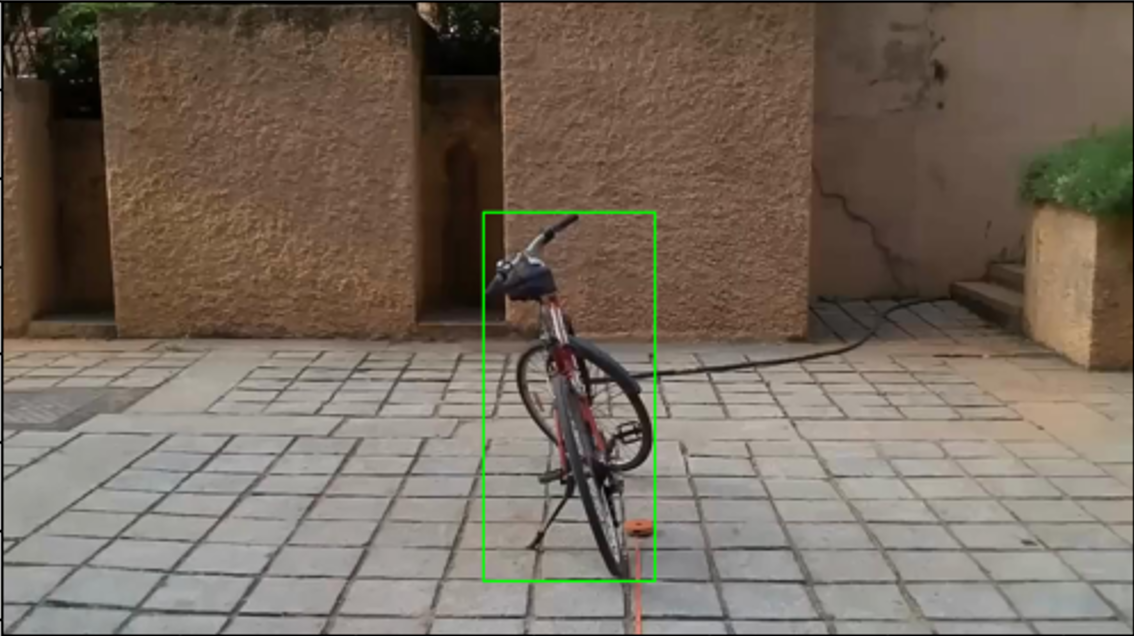}
    \label{fig:sample-cycle}
  }
  \subfloat[Car]{%
    \includegraphics[width=0.24\columnwidth]{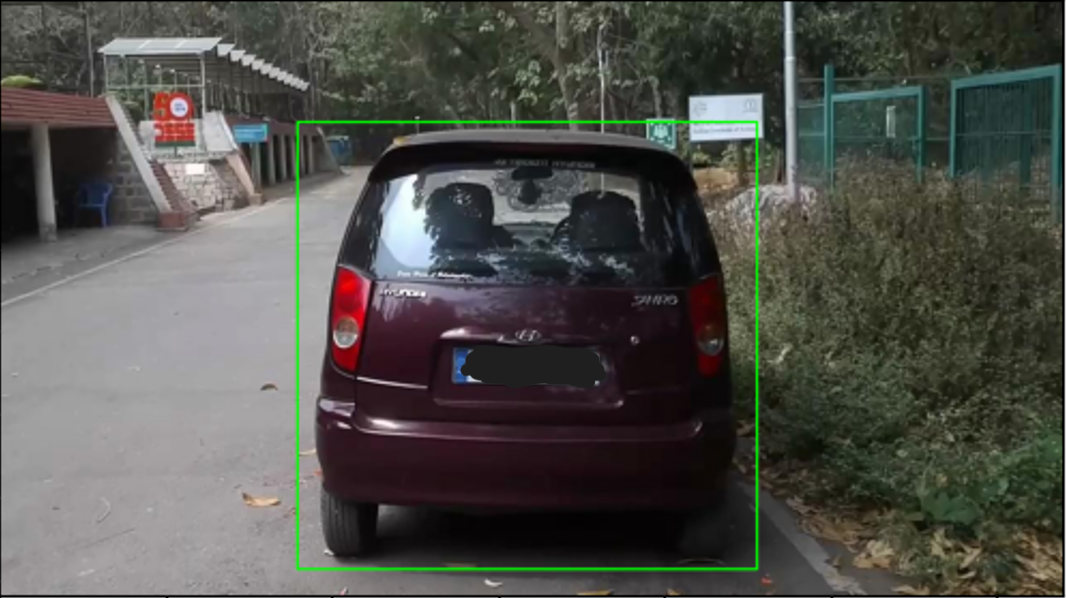}
   \label{fig:sample-car}
  }
    \subfloat[VIP with bystander obstacle]{%
   \includegraphics[width=0.24\columnwidth]{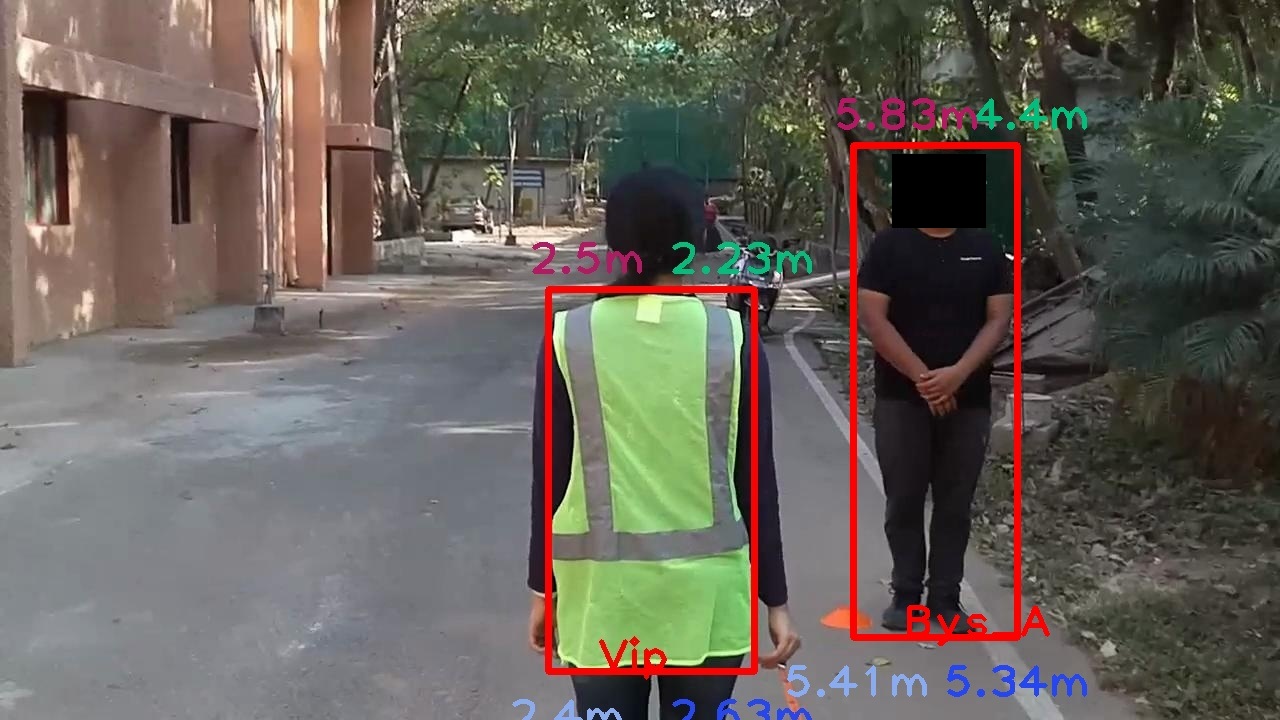}
    \label{fig:run1-f250}
  }
    \subfloat[VIP with bystanders and vehicle obstacles]{%
   \includegraphics[width=0.24\columnwidth]{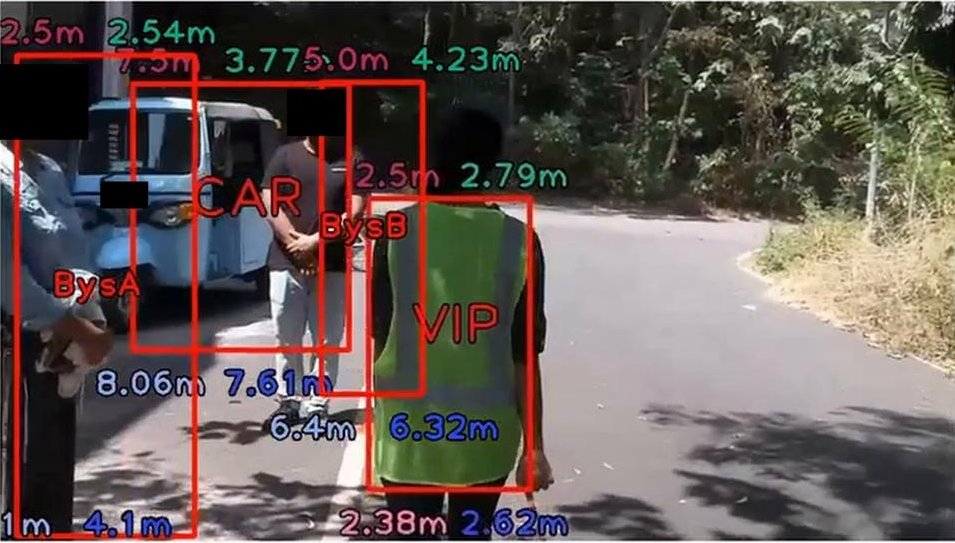}
    \label{fig:dese1-f250}
  }
\caption{Frames from videos of obstacles dataset collected for evaluation.}
\label{fig:objects-dataset}
\end{figure}

\subsubsection{Static VIP Dataset}\label{exp:data:vip}
We collect the videos of the VIP in $3$ different \textit{scene settings}, where different VIPs (student proxies) stand against different \textit{backgrounds} and lighting when the drone is also \textit{stationary}. The background/lighting conditions include: a plain light background in the morning (\texttt{AM\_Simple}), a green foliage in the morning (\texttt{AM\_Complex}), and a green foliage in the evening (\texttt{PM\_Complex}), as shown in Fig.~\ref{fig:scenes-description}. These scenarios cover a wide range of simple and complex outdoor scenes that we encounter in the real world. For each scene setting, we record a $10s$ video of the VIP at \textit{distances} of $2$--$4m$ from the drone, in steps of $0.5m$. With three different VIP proxies, we have $3~VIP \times 3~scene \times 5~distance$ combinations for a total of $450s$ of diverse videos for VIPs. We extract $\approx14.7k$~frames from these.

\subsubsection{Static and Dynamic Obstacles Dataset}\label{exp:data:obs}

\begin{table}[t]
\centering
\footnotesize
\caption{Summary of Distance Evaluation Video Dataset}
\setlength\tabcolsep{2pt}
% \begin{adjustbox}{width=\columnwidth}
\begin{tabular}{@{}l|c|c|l|r|r|r@{}}
\toprule
\textbf{Categories}       & \textbf{\makecell{Total \\Video\\ Length}} & \textbf{\makecell{Motion\\ in\\ Frame}} & \textbf{\makecell{Classes\\(Count)}} & \textbf{\makecell{\# of \\Video\\ Clips}} &  \textbf{\makecell{Proximity\\ from \\Drone}} & \textbf{\makecell{$\#$ Extracted\\ Frames\\ (FPS)}} \\ \midrule
\textbf{VIP}  & 490 s  & N  & VIP (3)  & 45 & 2-4 m   & 14719 (30)\\ \midrule
\multirow{4}{*}{\textbf{Obstacles}} & \multirow{4}{*}{400 s}      & \multirow{4}{*}{N} & Bicycle (2) & 10  & 2-4 m & 3300 (30) \\
& & & Car (2)  & 10 & 2-4 m & 3154 (30) \\
& & & Scooter (2) & 10  & 2-4 m  &  3146 (30)\\
& & & Bys. (2) & 10 & 2-4 m  &   3010 (30) \\ 
\midrule
\multirow{4}{*}{\textbf{\makecell{VIP with \\Obstacles\tablefootnote{These videos include the VIP in all frames as the drone followed them, but they are excluded from reporting since obstacles were the primary focus. Here, proximity from the drone refers to the initial distance, which decreases as the VIP moves closer to the obstacles.}}}} & \multirow{4}{*}{590 s} & \multirow{4}{*}{Y} & Bys. (3) & \multirow{4}{*}{4}   & 5-73 m & 195 (1) \\
& & & Car (1)& & 4-65 m  &  133 (1)\\
& & & Scooter (1) & & 17-62 m & 31 (1) \\
& & & Bicycle (1) & & 10-51 m & 64 (1)\\ 
\midrule
\multirow{2}{*}{\textbf{Adversarial\tablefootnote{VIP count is reported only for videos with exclusive VIP frames. In videos containing both bystander and VIP, only bystander frames are reported.}}} & \multirow{2}{*}{42 s} & \multirow{2}{*}{Y} & VIP (1)& 2  & 2.5m & 252 (30)\\
& & & Bys. (1)& 1 & 2.5m &  301 (30)\\                                     
\bottomrule
\end{tabular}
\label{table:video-dataset-summary}
\end{table}

To evaluate the distance estimation models for different outdoor obstacles, we first record videos with only a \textit{single stationary obstacle} present in the frame. We consider four common obstacle classes: bystander, cycle, car and scooter, we have two class instances for each, e.g., an SUV and a hatchback car, different bystanders, etc., with different backgrounds, and at distances of $2$--$4$~m from the drone in steps of $0.5$~m. Samples images are shown in Fig.~\ref{fig:sample-cycle} and~\ref{fig:sample-car}. We collect $\approx3000$ validation images for each class from a total of $400s$ of videos.

Additionally, we collect videos where the drone follows the VIP at a distance of $2.5m$, and different obstacle classes are encountered under diverse conditions, e.g., as shown in Fig.~\ref{fig:run1-f250} and~\ref{fig:dese1-f250}. As discussed, we also measure the ground truth distances to the obstacles from the drone along this trajectory. While these obstacles are stationary, their relative distances to the moving drone \textit{change dynamically} across the frames and are present in different backgrounds. We have 4 videos with a total duration of $590s$, from which obstacles data is extracted at $1$~FPS and VIP data at $30$~FPS. Bystanders, cars, scooters, and bicycles appeared in $195, 133, 31,$ and $64$ images, respectively.

\subsubsection{Adversarial Dataset}\label{exp:data:adv}

\begin{figure}[t]
    \centering
\subfloat[VIP bends]{
   \includegraphics[width=0.3\columnwidth]{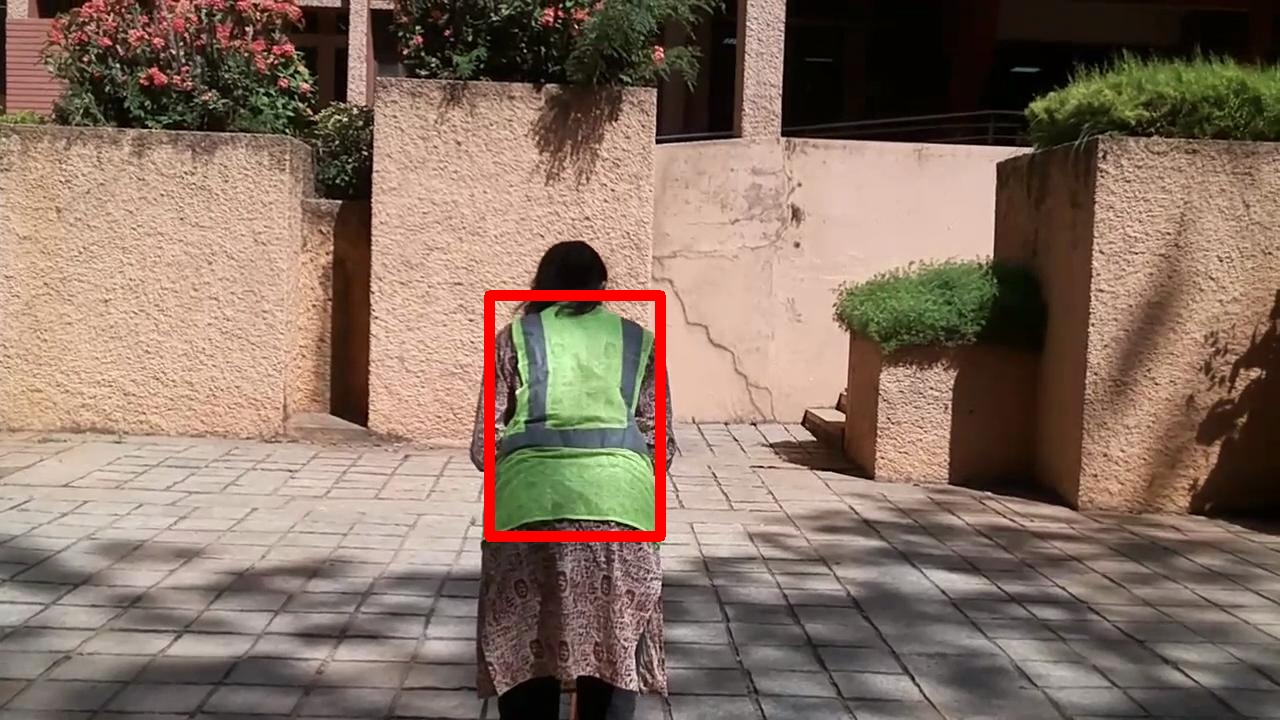}
    \label{fig:bend}
  }~~
  \subfloat[VIP sits]{
  \includegraphics[width=0.3\columnwidth]{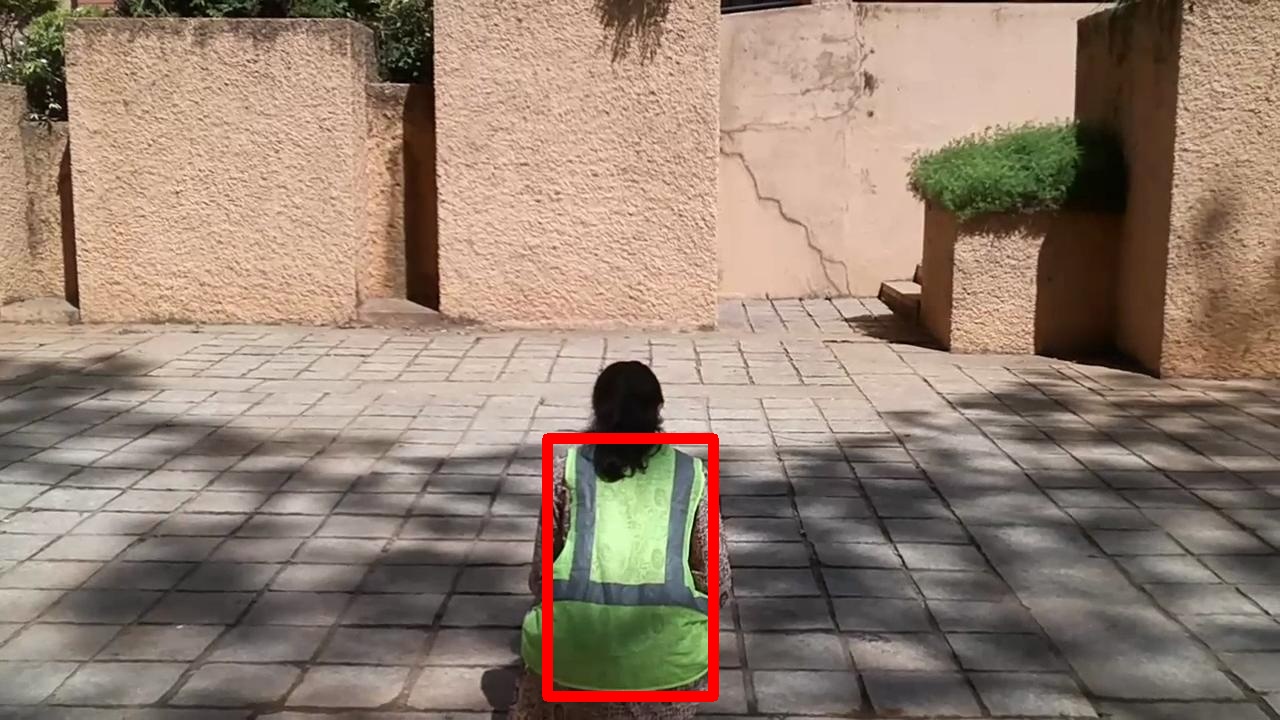}
    \label{fig:sit}
  }~~
  \subfloat[Truncated Bystander]{ 
  \includegraphics[width=0.3\columnwidth]{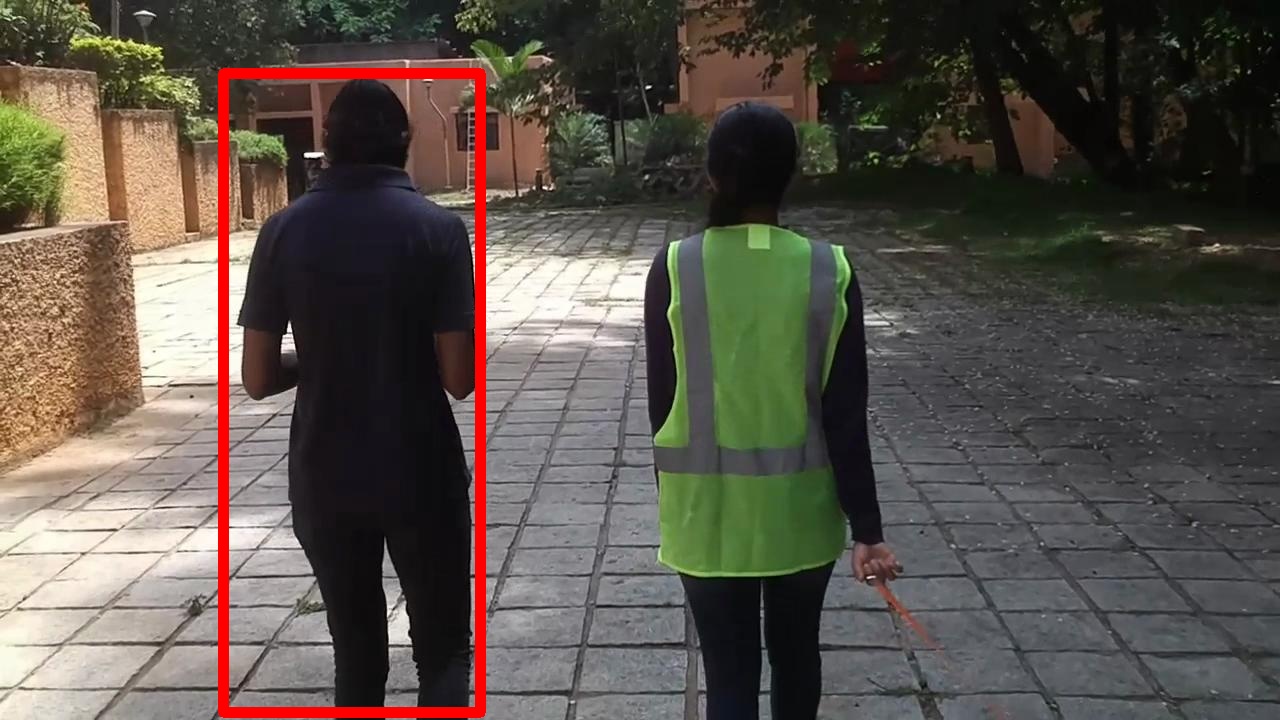}
   \label{fig:bystander-truncated} 
   }
    \caption{Adversarial images of VIP and bystander.}
    \label{fig:adversarialimages} 
\end{figure}

In addition, we also record $40s$ of \textit{adversarial scenarios} for the VIP and obstacles, e.g., where the VIP bends down, moves to the edge of the frame, etc. 
Sample adversarial scenarios for the VIP and a bystander include: (a) VIP bends (Fig.~\ref{fig:bend}), (b) VIP sits (Fig~\ref{fig:sit}), and (c) a bystander next to the VIP gets truncated (Fig.~\ref{fig:bystander-truncated}).

The effort in collecting these diverse and complex videos spanned a couple of months, and will be made public. A summary of the datasets is provided in Table.~\ref{table:video-dataset-summary}. 

%=============================================
\subsection{Calibration of Distance Estimation Models}\label{sec:eval:calib}
Next, we apply the calibration approaches proposed earlier to the datasets we have collected to decide the relevant coefficients. 

\subsubsection{Regression Model}
This model can only estimate distances to the VIP, not other objects. Since this is specific to a VIP, we calibrate it for each VIP by sampling $10$ frames from their \texttt{AM\_Simple} videos at distances of $\{2m,3m,4m\}$ from the drone, i.e., a few seconds of the VIP's videos at three distances is all that is required, imposing minimal one-time effort for the VIP--drone pair. Using this, we fit the linear regression Eqn.~\ref{eqn:regression} and get the coefficients $(a,b,c)$. For the three VIPs in our datasets, these coefficients are:  $(-2.42,-1.29,0.0043)$ for VIP1, $(-2.14,-1.77,0.0050)$ for VIP2, and $(-2.56,-2.10,0.0065)$ for VIP3.

\subsubsection{Geometric Models}
The Geometric and Geometric* models can estimate the distances to the VIP and to obstacles that are present among the \textit{80 known object classes} of YOLOv8 (medium). For calibration, we require the height $h_o$ for that object besides the focal length $f_c$ of the camera in Eqn.~\ref{eqn:final}. For \textit{Geometric}, this is the expected (average) height for the object class, and for \textit{Geometric*}, it is the actual height for that specific object instance. The expected heights are sourced from public information, and is a one-time effort, while the actual height for an object is impractical to get in practice; we manually measure it for the datasets we collect and it serves as an ``ideal'' case for this model.
In our datasets, the heights of the obstacle class instances are: Bystanders (expected=$1.65m$, actual=$\{1.75m,1.76m\}$); Scooters (expected=$1.12m$, actual=$\{1.14m,1.25m\}$); Bicycles (expected=$0.97m$, actual=$\{0.95m,1.18m\}$); Cars (expected=$1.7m$, \\actual=$\{1.56m,1.38m\}$.

\subsubsection{\neo Model}
\textit{Static calibration} helps decide the \textit{scale ($m$)} and \textit{shift ($s$)} values for MonoDepth2 offline (\S~\ref{sec:depth-params-estimate}) and is used for all subsequent scenes by \neo. While we calibrate using the VIP images similar to Regression, we reuse these coefficients to decide the actual depths to all objects in the videos -- this re-usability is a key benefits of \neo, unlike Regression. Here, we select a pair of distances to the VIP, sample $10$ images at these two distances from the \texttt{AM\_Simple} videos, and use these to fit $m$ and $s$ in Eqn.~\ref{eqn:depth-map}. From our experiments using distances of $\{2m,2.5m,3m,3.5m,4m\}$, we find that the distance-pair of $(2.5m,4m)$ offers the lowest average median errors across diverse samples. These two distances are far apart and able to capture the variability in the VIP distances. So, using $10$ images at these two distances, we calibrate the coefficients for each of the three VIPs, and find their $(m,s)$ values to be $(1169, 124.2)$, $(1685, 54.9)$, and $(1105, 118.5)$, respectively.

In \textit{dynamic recalibration}, we update the coefficients at on-the-fly if the estimation error to the VIP using \neo relative to the Regression model exceeds $\tau=30cm$ over a window of $w=5$ frames; we evaluate different detection window sizes of $w=\{5,10,15\}$ and find $5$ to be the best. If the error exceeds $\tau$, we recalculate $m$ and $s$ over a sliding window of VIP frames over $w'=5$~secs, with $\alpha=0.75$ used as the weighing ratio; this was the best after evaluating $\alpha = 0.25$--$0.75$.

These parameters need to be \textit{calibrated once} for a given drone camera. For \textit{Regression} and \neo, the camera can affect the bounding box reported by Yolo, used for by these models. For Geometric, the focal length of the camera needs to be computed or used from the camera datasheet.

%% #############################################################################################
%% #############################################################################################
\section{Evaluation Results}\label{sec:evaluation}

We use the above calibrated distance estimation models, regression, geometric and \neo, to evaluate the accuracy for the diverse set of validation videos that we have collected. We report and analyze these results here~\footnote{A \href{https://figshare.com/s/1df990c276c326067c7c}{supplementary video (https://figshare.com/s/1df990c276c326067c7c)} helps visualize our evaluation results.}. Additionally, we present the results from the existing State-of-the-Art (SOTA) methods, Monodepth2 and \midas + ZoeDepth; the comparison with SOTA is concise as they consistently perform worse than \neo and our proposed baselines.

By default, we report the the signed error (in cm) for these evaluations, where $Error = (true~distance - predicted~distance)$. Positive errors indicate an \textit{underestimate} of the distance from the drone, while negative errors represent an \textit{overestimate}. We use $\pm30cm$ error as a nominal safety tolerance for obstacle avoidance, with overestimates preferred over underestimates for conservative behavior.

%============================================
\subsection{Distance Estimation for VIP}

\begin{figure}[t]
    \centering
    \begin{minipage}{0.65\columnwidth}
        \includegraphics[width=0.95\columnwidth]{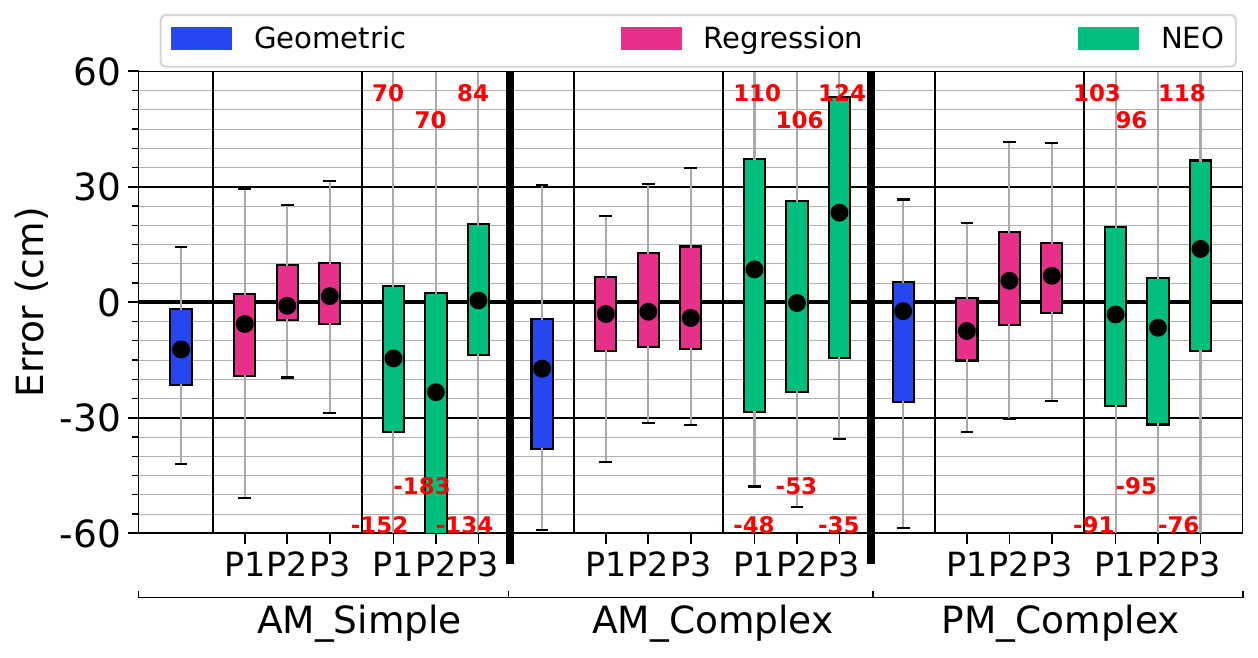}
        \caption{Accuracy of distance estimation for static VIP for simple and complex scenes.} 
    \label{fig:comparison-over-vip}
    \end{minipage}
    \qquad
    \begin{minipage}{0.25\columnwidth}
        \subfloat[Vest cropped at bottom]{\includegraphics[width=0.9\columnwidth]{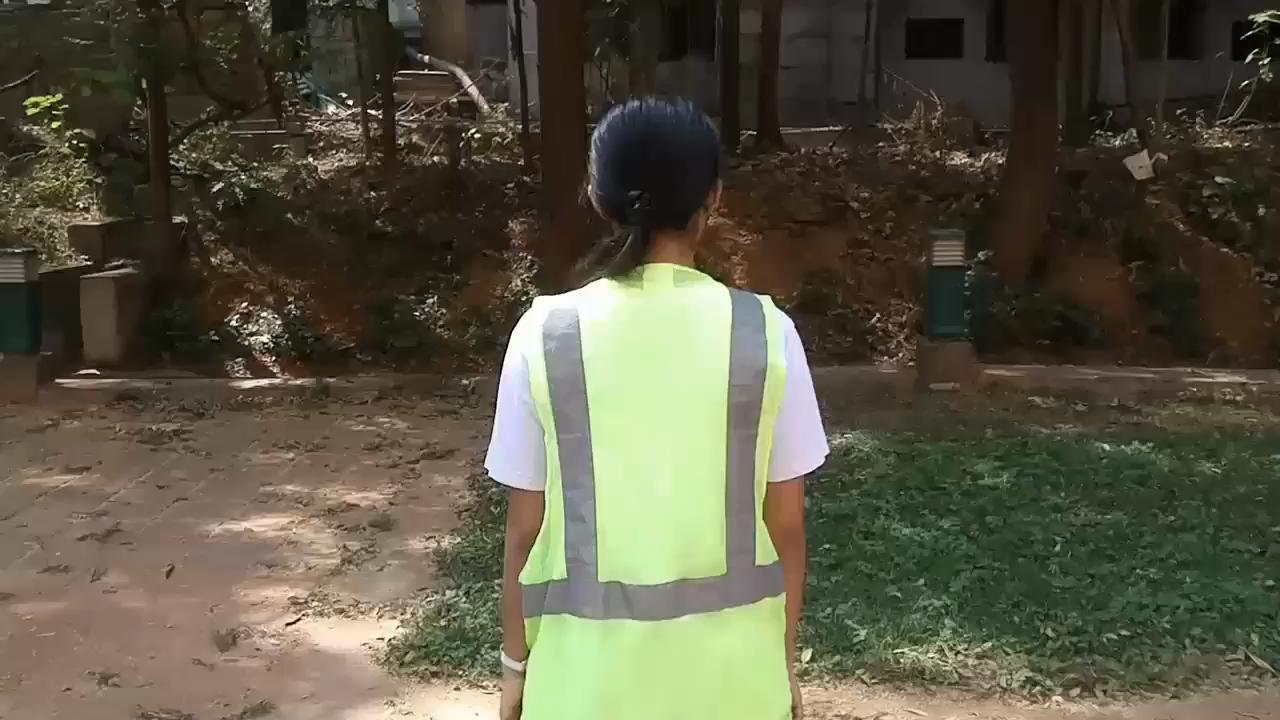}\label{fig:b-green-morn-f0}}\\
        \subfloat[Shiny surface of a Car]{\includegraphics[width=0.9\columnwidth]{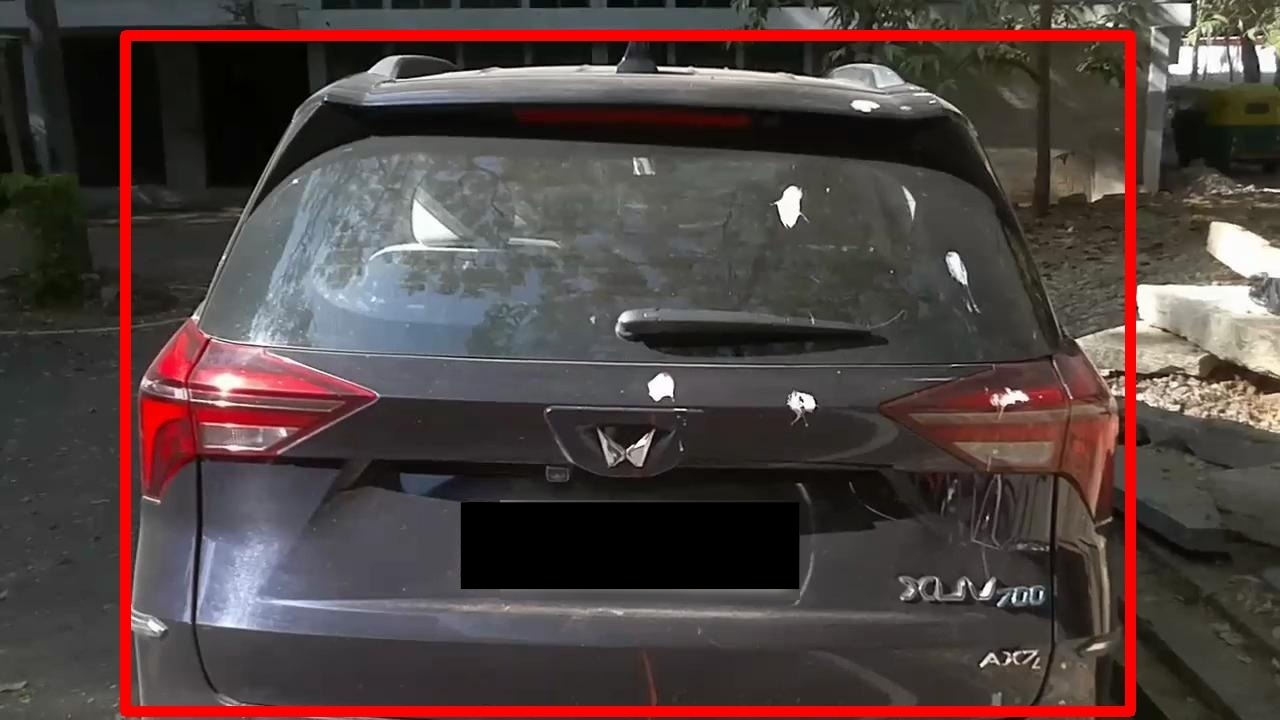}
   \label{fig:car-blurred}}
   \caption{Outlier images from VIP and Obstacles dataset.}
    \label{fig:distance-estimation-dynamic-objects}
    \end{minipage}
\end{figure} 

First, we evaluate the accuracy and robustness of distance estimates to a \textit{static VIP}. For this, we use the stationary VIP video datasets described in \S~\ref{exp:data:vip}, with varying VIPs, distances and backgrounds/lighting conditions.

Figure~\ref{fig:comparison-over-vip} reports the error for the distance estimate to the VIP given by the \textit{Geometric, Regression} and \textit{\neo} models, relative to the actual ground truth distance. 
The box and whiskers plot show the error distribution across all three VIP video frames ($4.8k$) for each model, with the box showing Q1--Q3 quartiles, the marker indicating median (Q2), and the whiskers the min and max. For Regression and \neo, the models are calibrated using a specific VIP (P1, P2, P3), shown on the X axis, but validated across datasets from all VIPs to understand their generalizability. In contrast, Geometric uses the same expected height for all VIPs since they belong to the same object class. 

Our \textit{Regression} model (pink bars) provides the best distance estimation for VIP with a median error range of only $-8cm$ to $7cm$, irrespective of the VIP for which it was originally calibrated on. These models also have a tighter distribution. \textit{Geometric} performs the next best with a median error between $-18cm$ to $-2cm$, with the peak errors going up to $-39cm$. The higher error is caused by the hazard vest being partially cropped from the image when the drone is close to the VIP (e.g. $2m$ in Fig.~\ref{fig:b-green-morn-f0}). This causes the height of the bounding box to reduce and results in an overestimate of the distance. \textit{\neo} has a maximum median error of $\pm 25cm$, but a higher variability, especially in the \texttt{AM\_Complex} scene due to the bright sunlight reflecting off of the background and the VIP's vest (e.g., Fig.~\ref{fig:am-complex}). This reduces during the evening, when the brightness is uniform for all pixels of the vest (e.g., Fig.~\ref{fig:pm-complex}).

Finally, we evaluate the accuracy of SOTA methods. Monodepth2 achieves an absolute median error of $141cm$, $174cm$, and $162cm$ for \texttt{AM\_Complex}, \texttt{AM\_Simple}, and \texttt{PM\_Complex}, while \midas + ZoeDepth reports errors of $154cm$, $174cm$, and $196cm$ for the same datasets. These numbers are excessively high and render the methods unsuitable for practical scenarios.

\textbf{Discussion.~} The median errors for all strategies are within $\pm30cm$, our safety tolerance. Also, Regression and \neo, despite being calibrated on one VIP, generalize well across different VIPs, showing comparable errors for P1--P3. However, the distance estimate to the VIP from the drone needs to be more accurate and robust than the $\pm30cm$ safety margin for other obstacles. This estimate is actively used for drone and VIP navigation, and will compound the distance errors to other objects. The Regression model has a median error of $\pm 8cm$ and the tightest distribution, with Q1--Q3 being $\pm 20cm$. While not shown, Regression also has the best accuracy even when the VIP and drone are moving, with a peak error of $\leq 15cm$, while Geometric and \neo have errors of $\approx 30cm$. So, we adopt Regression as our \textit{default model} for VIP distance estimation in later experiments.

%===================================================
\subsection{Distance Estimation for Static Obstacles} 

\begin{figure}
    \centering
    \includegraphics[width=0.5\columnwidth]{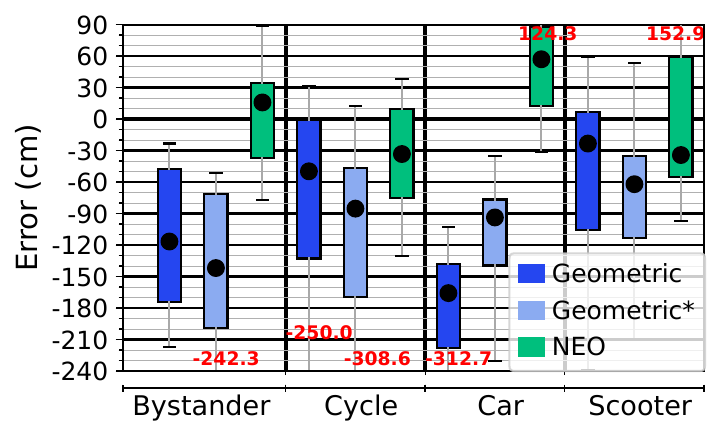}
    \caption{Accuracy of distance estimation for static obstacles.}
    \label{fig:static-comparison-over-objects} 
\end{figure}

Next, we validate the Geometric and \neo models to estimate distances to obstacles in the vicinity of the VIP; Regression works only for the VIP and is omitted. Here, we consider static scenes, i.e., neither the VIP nor the obstacles are moving; dynamic scenes are considered in the next section. 
We evaluate the models on the $\approx3000$ validation images for bystanders, cycles, cars and scooters with different backgrounds and distances from \S~\ref{exp:data:obs} and sample images in Figs.~\ref{fig:sample-car} and~\ref{fig:sample-cycle}.

Fig.~\ref{fig:static-comparison-over-objects} reports the distance error for these static obstacles. For obstacle avoidance, \textit{under estimates (positive errors) are better than over estimates (negative errors)} since it is better to take corrective measures assuming an object is closer than reality, than assume it is farther away and potentially collide, i.e., high \textit{recall} is more important than high \textit{precision}.

\neo~performs consistently better than the \textit{Geometric} methods and exhibits a peak median error of $56cm$ for cars and falls within a median of $35cm$ for other obstacles. The pixel normalization used by \neo over the depth map image (\S~\ref{sec:score-normalization}) ensures that it is robust to scenes where the images of the obstacles are cropped, unlike the Geometric method. \neo, however, is susceptible to reflections from shiny surfaces (as seen before for the reflective vest in Fig.~\ref{fig:am-complex}), which it also encounters in cars (Fig.~\ref{fig:car-blurred}) and causes a positive error. This increases with an increase in the actual distance.

Geometric performs worse as it is dependent on the entire image of the obstacle being visible. This often fails when the objects are nearby, \textit{precisely when accurate estimates are important}. For an object, this ratio equality should hold: $\frac{{h_o}}{h_{b}} = \frac{d}{f}$. For the focal length of $1592$~pixels for the Tello camera, this ratio translates to values of $\{0.12, 0.15, 0.19, 0.22, 0.25\}$ for distances of $\{2,2.5,3,3.5,4\}m$, respectively. However, for obstacles closer to the drone (and hence the VIP), cropping of the image and its bounding box causes this ratio to not hold and the shorter (visible) height leads to a distance overestimate. E.g., cycles at closer distances of $3$--$2m$ have errors grow from $26$--$124\%$. Errors also amplify for taller objects, e.g., bystanders are cropped even at $3m$ and have $85\%$ error. 

Geometric also diverges from the idealized \textit{Geometric*} when the actual height of the object is taller than the expected height, or due to truncation. E.g., the median error increases by $39cm$ for Geometric* compared to Geometric for scooters as its expected height is $112cm$ while the actual heights are $114cm$ and $125cm$. In contrast, the error decreases by $72cm$ for a car, where the average height is $170cm$ and actual is $138cm$.

Lastly, the results from SOTA methods for static obstacles were compared in \S~\ref{sec:comparison-with-sota}. As we saw, \neo achieves up to $5.3\times$ lower error than Geo*, $14.3\times$ than \midas, and $14.6\times$ than Monodepth2.

%===================================================
\subsection{Distance Estimation for Dynamic Obstacles}\label{sec:exp:analysis-dynamic-obstacles}
\begin{figure}
% \vspace{-0.1in}
\centering%~%
  \subfloat[Car]{\includegraphics[width=0.24\columnwidth]{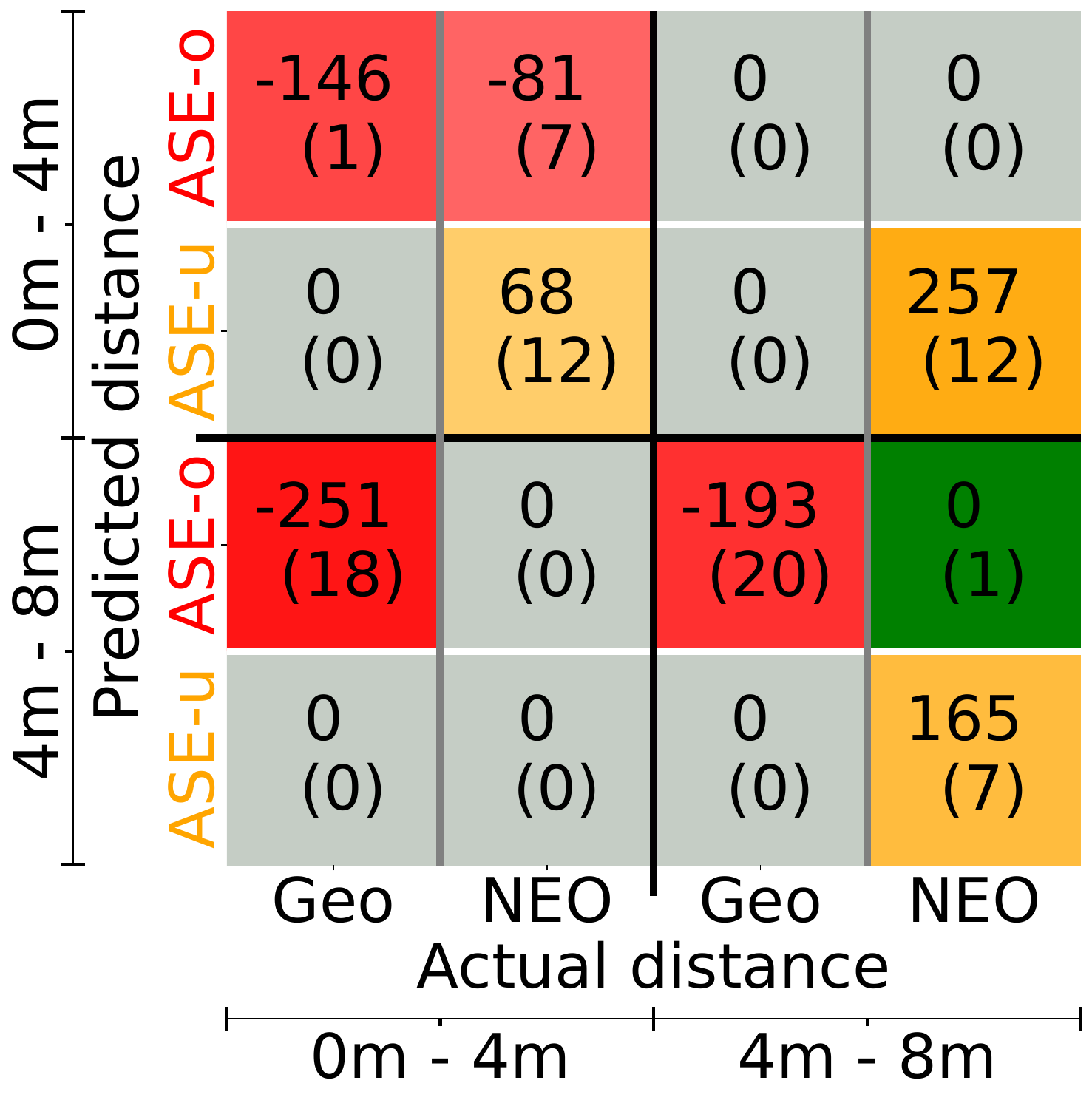}
    \label{fig:car}
  }
\subfloat[Cycle]{
   \includegraphics[width=0.24\columnwidth]{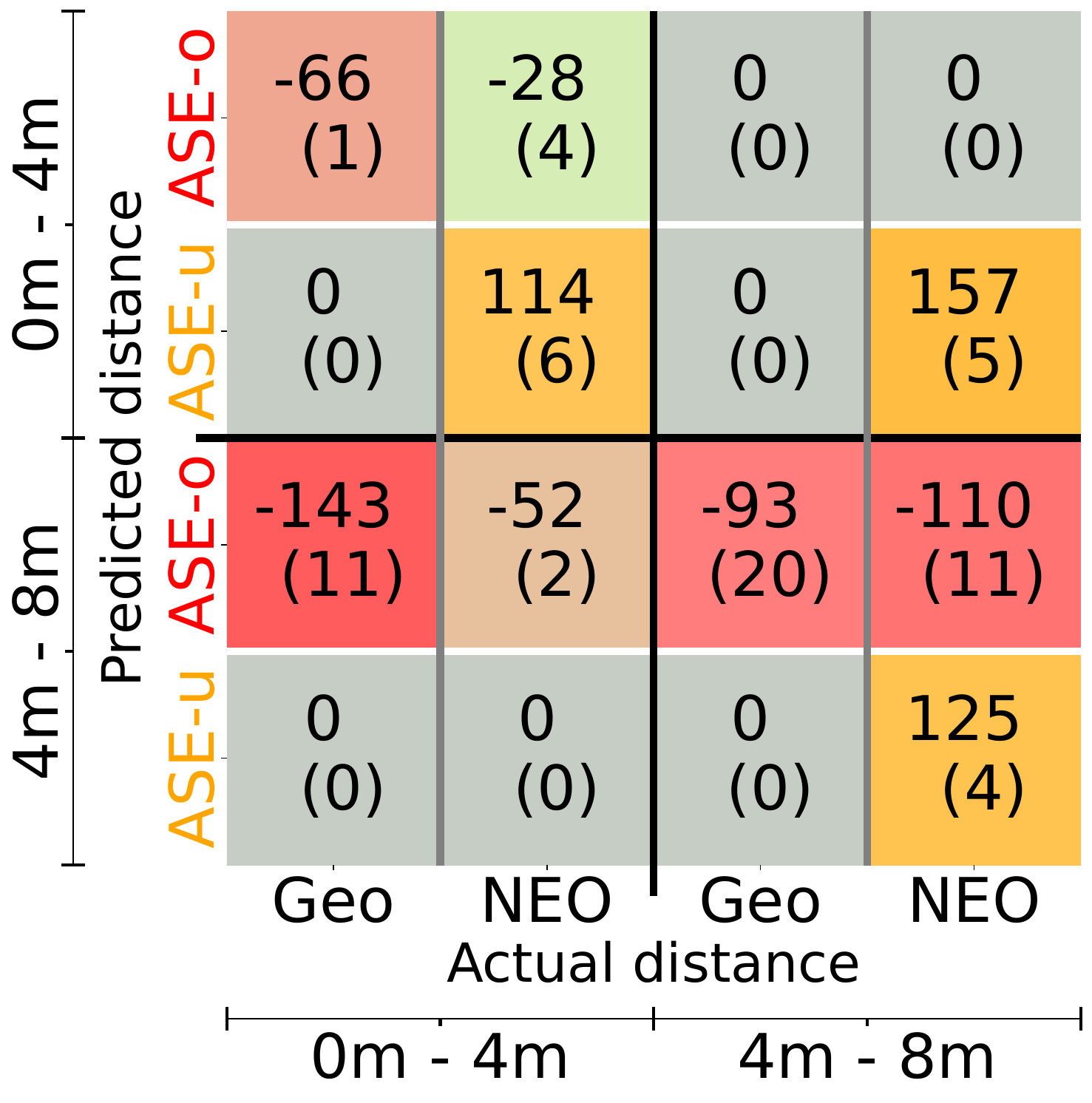}
    \label{fig:cycle}
  }
  \subfloat[Scooter]{
   \includegraphics[width=0.24\columnwidth]{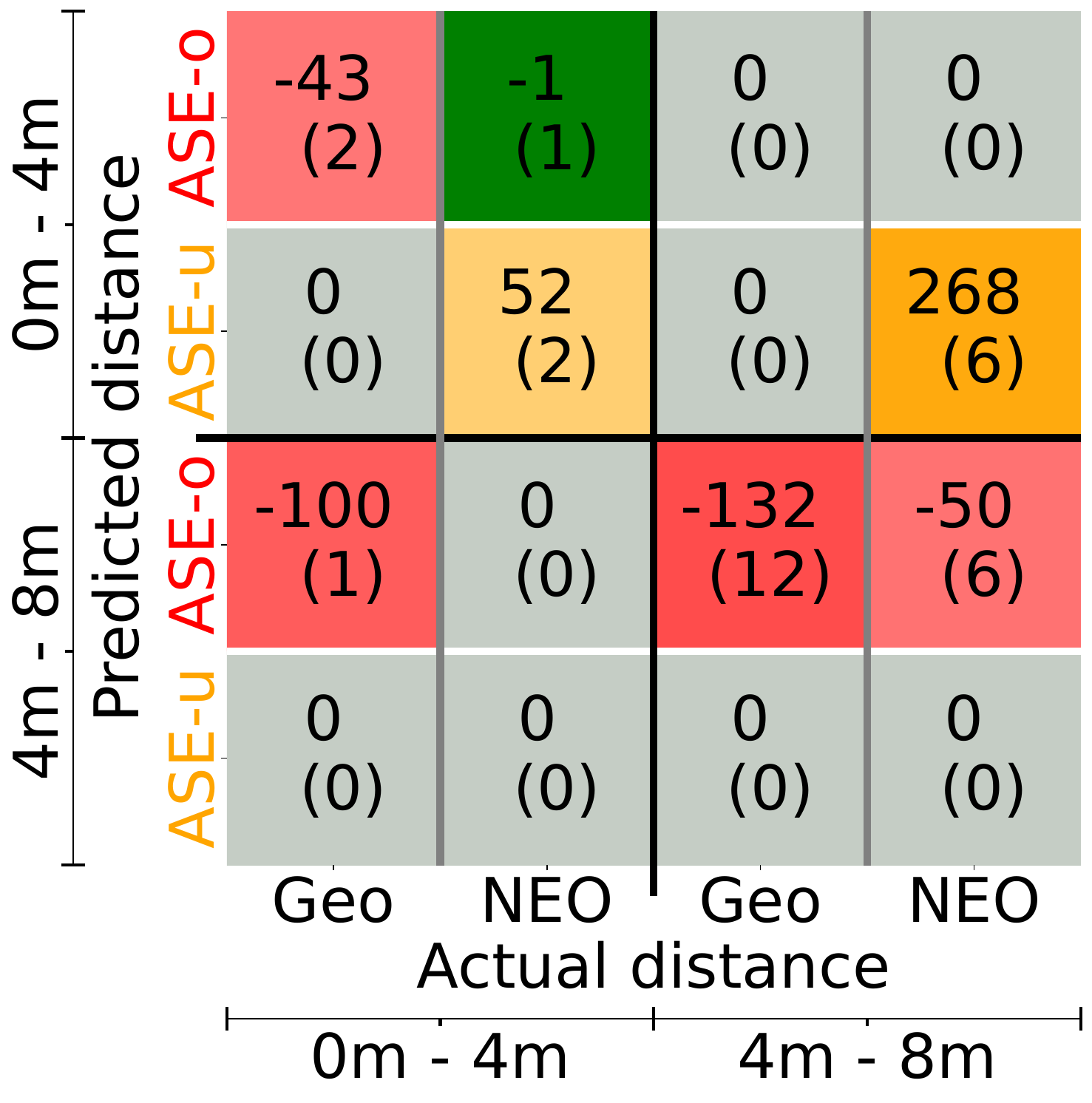}
    \label{fig:scooter}
  }
  \subfloat[Bystander]{ \includegraphics[width=0.24\columnwidth]{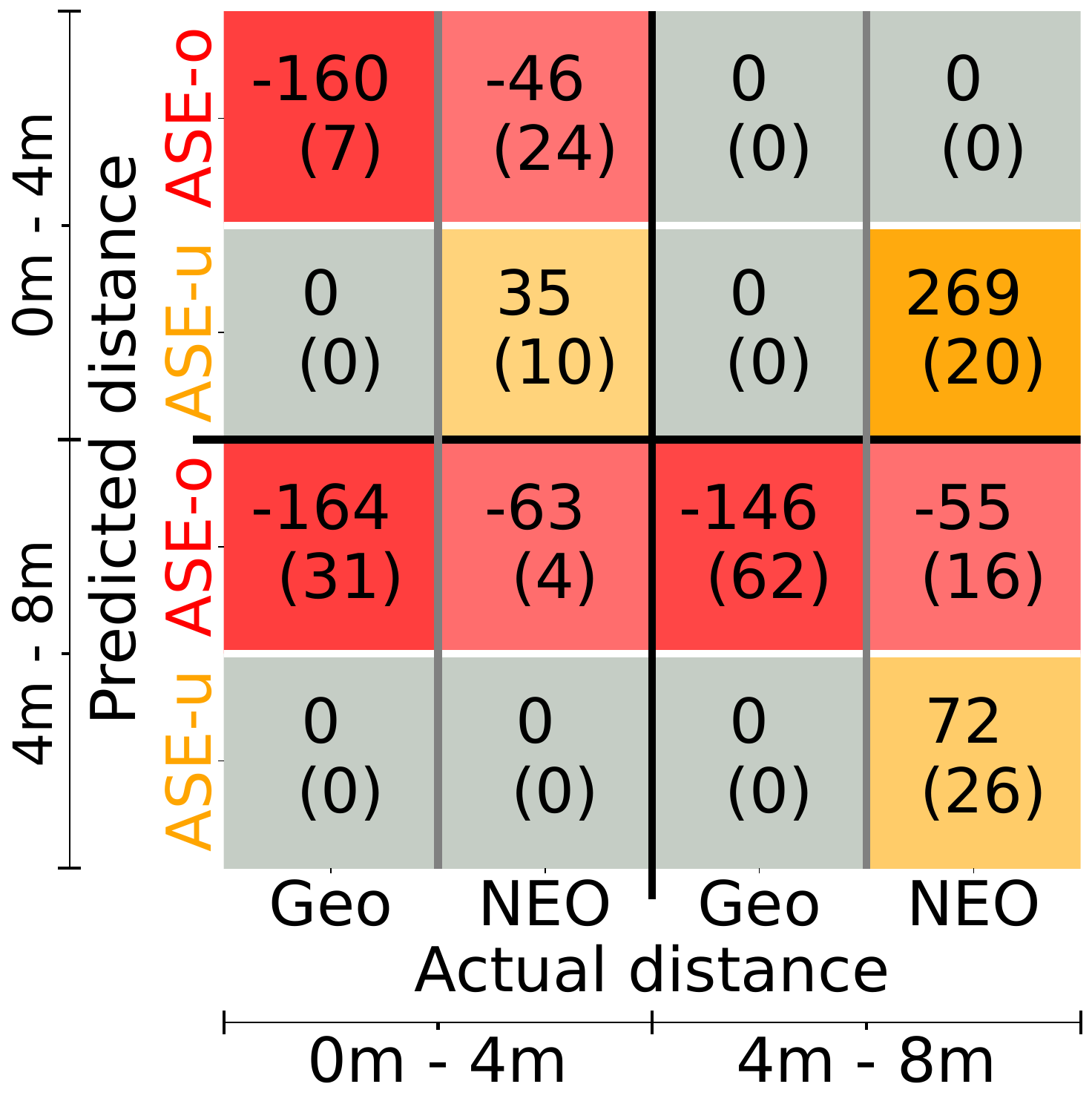}
   \label{fig:bystander}
  }
\caption{Matrix showing over- and under-estimation errors (ASE in $cm$) for \neo and Geometric (inner quadrants), and at distances of $\leq 4m$ and $>4m$ (outer quadrants), for obstacle types present in dynamic scenes. The \# of frames in each cell is shown in parentheses and colors indicate extent of error.}
\label{fig:heatmap-confusion-matrix}
\end{figure}

Next, we validate our models for realistic scenarios where the VIP is walking with the drone following, and various stationary obstacles are moving relative to the drone within the scene.
We use the dataset from \S~\ref{exp:data:obs} with $133,64,31$ and $195$ video frames of car, cycle, scooter and bystander, respectively, with sample images shown in Figs.~\ref{fig:dese1-f250} and~\ref{fig:run1-f250}. The ground truth distance between the drone's initial position and these obstacles ranges from $4$--$73m$. This distance decreases as the drone follows the VIP and gets closer to the objects. We limit our analysis to objects within $8m$ from the drone, and use the dynamic variant of \neo with recalibration.

We categorize the images into: \textit{(i)} objects $\leq 4m$ to the drone, which can pose an imminent danger to the VIP, and \textit{(ii)} objects $4$--$8m$ from the drone with lower risk. These are represented using a matrix for each obstacle type in Fig.~\ref{fig:heatmap-confusion-matrix} whose outer quadrants show the ASE for the objects at closer and farther distances. E.g., Q2 (top left quadrant) show errors for samples where both real and predicted distances are $\leq 4m$, while Q3 (bottom left) shows errors when real distance is $\leq4m$ but predicted is $>4m$. For each obstacle class, within a quadrant, we report the \textit{Average Signed Error (ASE) =} $\frac{\sum{(true~distance - predicted~distance)}}{\#~of~frames}$, in $cm$, with \textit{ASE-o} and \textit{ASE-u} separately for Geometric and \neo, indicating over and under prediction errors.
The number of frames in each category is shown in parenthesis for context, e.g., some may have $0$ frames in that bucket (grey cells). Cells with errors $\leq30cm$ are in green and perform the best. As before, we prefer lower ASE-u under-predictions (yellow) to ASE-u over-predictions (red), and also better predictions for closer objects than farther, i.e., errors in Q3 should be the least, followed by Q2, Q1 and Q4. 

\begin{figure}[t]
    \centering
  \subfloat[Q1: Underestimate of far objects]{%
  \includegraphics[width=0.27\columnwidth]{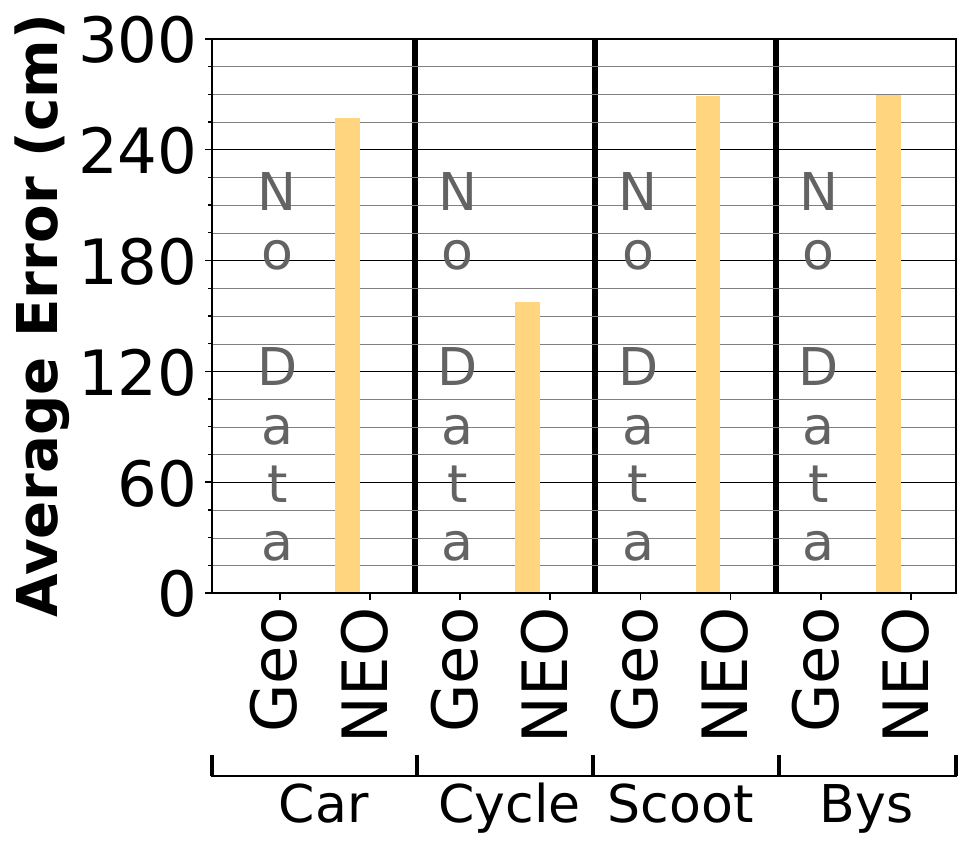}
    \label{fig:quad1}
  }\quad \quad
\subfloat[Q2: Underestimate of near objects by \neo]{%
   \includegraphics[width=0.25\columnwidth]{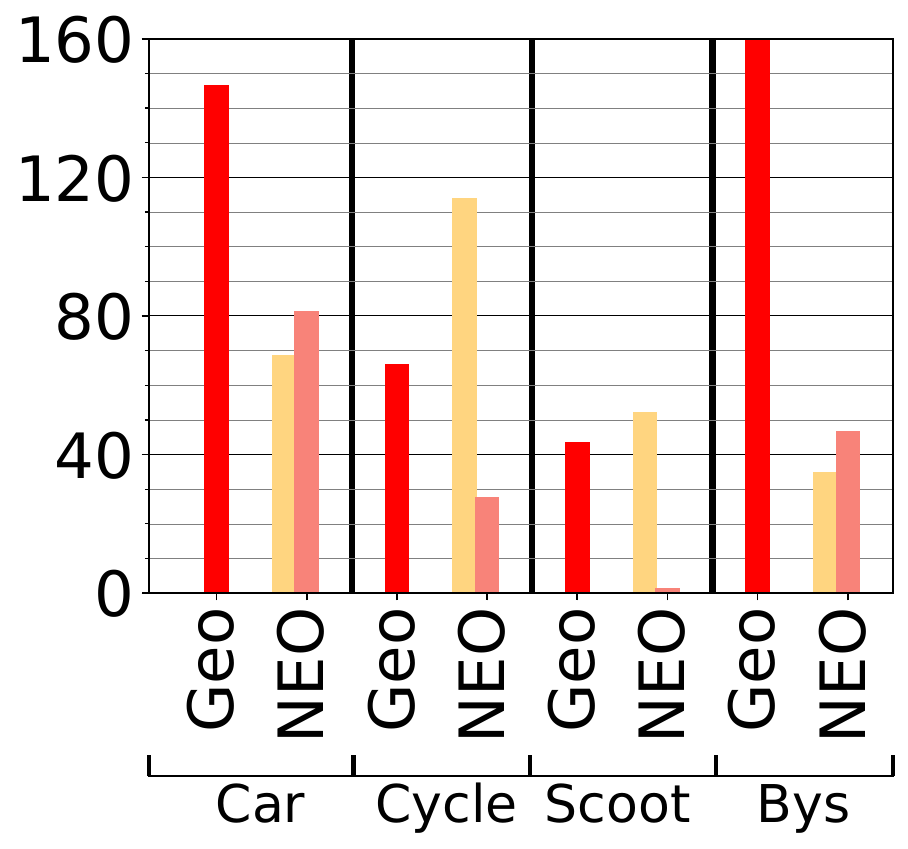}
    \label{fig:quad2}
  }\quad \quad
  \subfloat[Q3: Overestimate of near objects]{%
  \includegraphics[width=0.25\columnwidth]{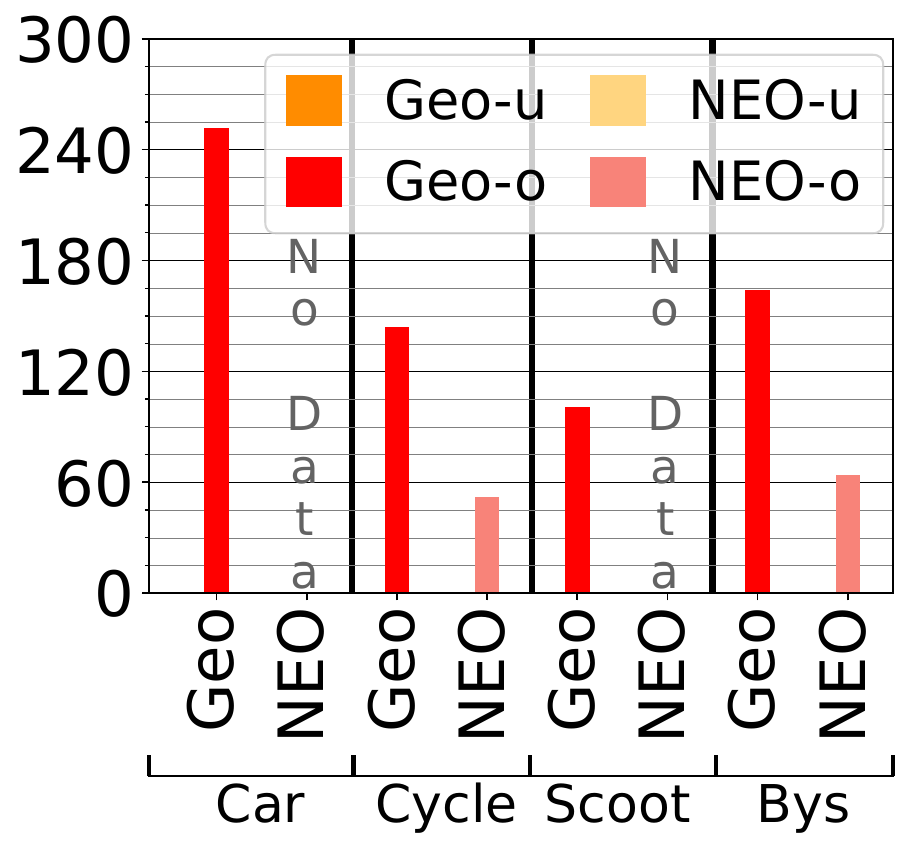}
   \label{fig:quad3} 
   }
    \caption{Per-quadrant analysis of overestimates (-o) and underestimates (-u).}
    \label{fig:quadrant-analysis} 
\end{figure}    

For \textit{cars} (Fig.~\ref{fig:car}), Geometric overestimates the distances for all $39$ images while \neo overestimates them only for $7$ of $39$. For objects closer than $4m$, \neo has ASE-u of $68cm$ and ASE-o of $81cm$, which is $65cm$ better than Geometric. Geometric also overestimates the distances by $251cm$ (false negative), which can be dangerous as the objects are closer than predicted. On the other hand, \neo under-predicts for farther objects, estimating them to be nearer (false positive), which is relatively acceptable.
For \textit{scooters} (Fig.~\ref{fig:scooter}), we see a similar trend; Geometric overestimates the distances for all $15$ images, while \neo overestimates only for $7$ out of $15$ images. For objects within $4m$, \neo achieves an ASE-o of just $1cm$, demonstrating significantly better accuracy. This extends to other classes like \textit{cycle} (Fig.~\ref{fig:cycle}) and \textit{bystanders} (Fig.~\ref{fig:bystander}). Here, Geometric overestimates distances for \textit{all} images while \neo limits this to $52\%$ of images. \neo also reduces absolute errors by up to $91cm$ for cycles and $114cm$ for bystanders when within $4m$, highlighting its benefits for closer objects.

This is further validated in Fig.~\ref{fig:quadrant-analysis}, where we summarize the over and underestimation errors for the most relevant quadrants, Q1, Q2 and Q3. If no instances of overestimation or underestimation are observed for a particular object using a given method, we report this as \textit{``No Data''}. Across all obstacle classes, \neo demonstrates consistently lower errors or tends to underestimate distances (yellow shade). This is clearly evident in Q1 (Fig.~\ref{fig:quad1}), where \neo underestimates distances for all classes with a ground truth distance that is farther off, at $4$--$8m$. 
We focus on Q2 (Fig.~\ref{fig:quad2}), which is key for our application as it has on objects with true-distance $<4m$ of the VIP but the underestimated distances are also within $4m$. Here, \neo shows a significant improvement. When it overestimates (\neo-o), the magnitude of its error is substantially lower compared to Geometric's overestimation errors (Geo-o). E.g., for bystanders, \neo's error is $\approx70\%$ lesser than Geo-o, ensuring better accuracy and safety during close-range navigation. Lastly, in Q3 (Fig.~\ref{fig:quad3}), \neo overestimates distances of near objects only for cycles and bystanders classes. Even here, the magnitude of error is much smaller compared to the Geometric model, further highlighting \neo's reliability.

Lastly, we report that the median errors from SOTA methods range between $327$--$390cm$ for \midas + ZoeDepth and $98$--$264cm$ for Monodepth2, which are substantially worse than \neo. 

\textbf{Discussion.} Overall, the results highlight \neo's better  performance compared to Geometric across multiple obstacle classes, particularly in critical scenarios where objects are within $4m$ of the VIP. By significantly reducing the frequency and magnitude of overestimation errors and maintaining consistently lower error values across all quadrants, \neo ensures safer and more reliable obstacle detection. These make \neo highly suitable for VIP use cases as it can enhance navigation safety and dependable obstacle avoidance, even in dynamic environments.

\begin{figure}[t!]
    \centering
    \begin{minipage}{0.58\columnwidth}
\centering%~%
  \subfloat[Car]{%
  \includegraphics[width=0.5\columnwidth]{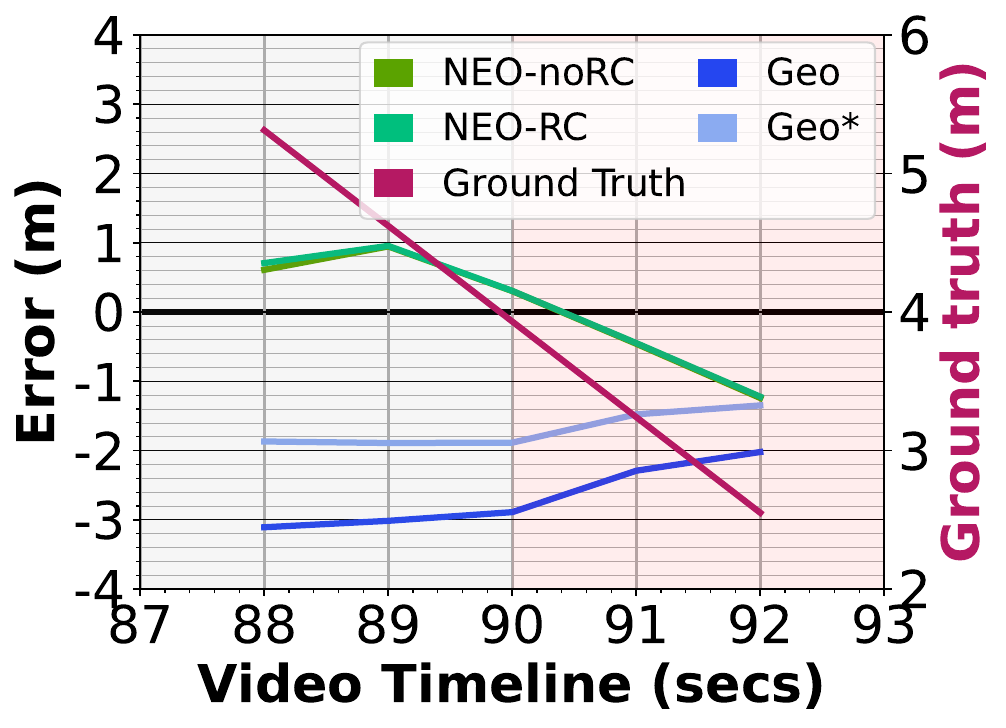}%
    \label{fig:temporalcar}%
    }
\subfloat[Cycle]{%
   \includegraphics[width=0.5\columnwidth]{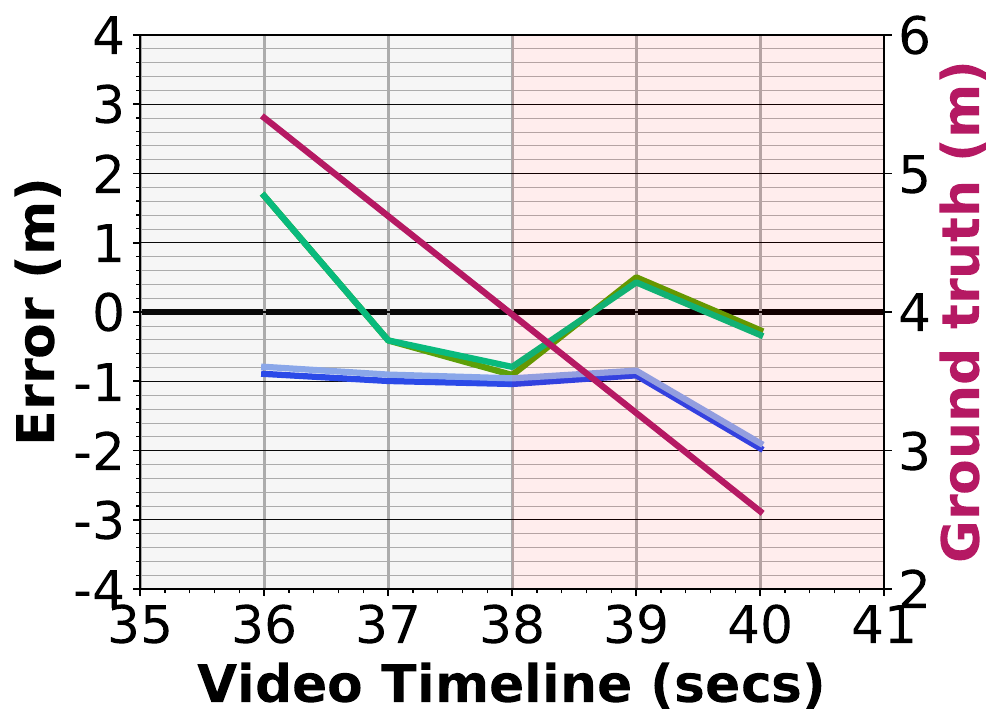}%
    \label{fig:temporalcycle}%
  }\\
  \vspace{-0.1in}
    \subfloat[Scooter]{%
   \includegraphics[width=0.5\columnwidth]{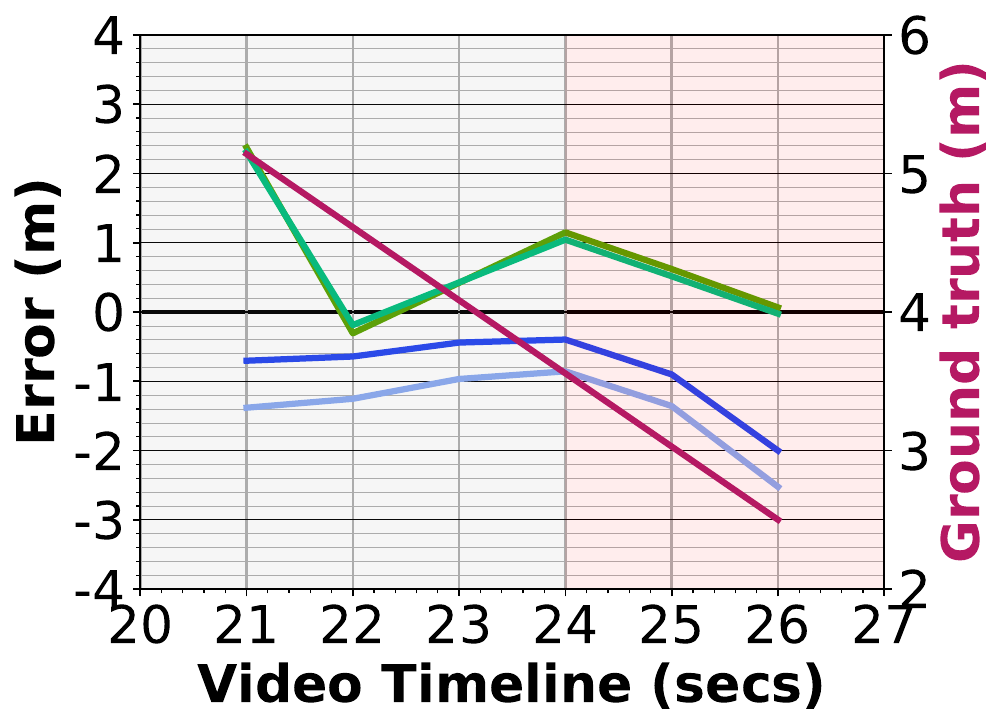}%
    \label{fig:temporalscooter}%
  }
  \subfloat[Bystander]{%
  \includegraphics[width=0.5\columnwidth]{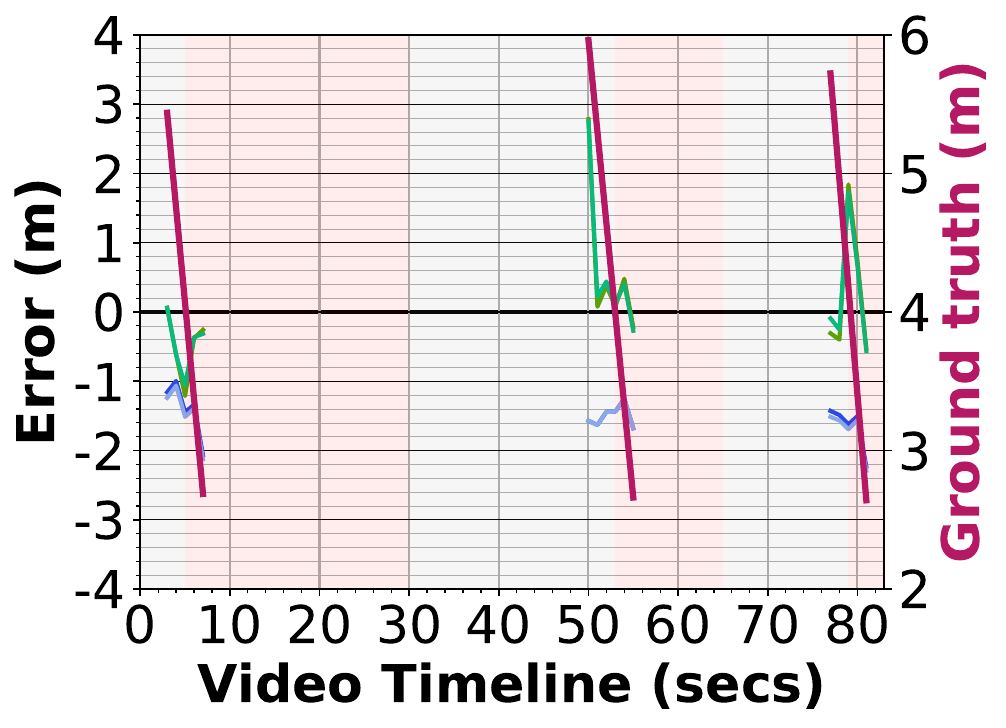}%
   \label{fig:temporalbystander}%
  }%
\caption{Error over time for objects, as VIP is moving. Gray background indicates object $>4m$ from drone while red is $<4m$.}
\label{fig:delta-error}
    \end{minipage}
    \quad
    \begin{minipage}{0.38\columnwidth}
\centering
    \includegraphics[width=1\columnwidth]{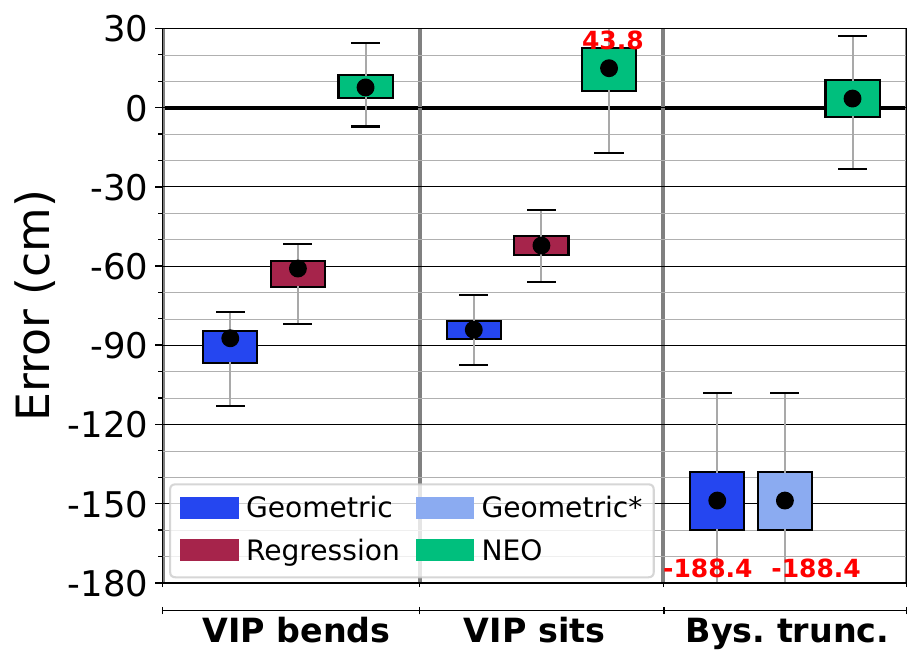}
    \caption{\neo performs better for adversarial scenarios.}
    \label{fig:adversarial}
    \end{minipage}
\end{figure} 

%===================================================
\subsection{Time-varying Errors for Dynamic Obstacles}

Now we examine errors across time in dynamic scenarios by analyzing one of the four dynamic videos in \S~\ref{exp:data:obs}. In this $206s$ long video, the VIP and drone are moving, and the obstacles are stationary but moving relative to the drone camera in the scene; hence the same obstacle appears at different points in time and at different distances. The temporal occurrence of obstacles are: bystander$_{1}$ at $t=5s$, scooter at $21s$, cycle at $36s$, bystander$_{2}$ at $49s$, bystander$_{3}$ at $76s$ and car at $88s$.

We evaluate the distance estimation methods: \textit{Geometric} (Geo), \textit{Geometric*} (Geo*), \neo without recalibration (\neo-noRC), and \neo with dynamic recalibration (\neo-RC), against the ground truth data.
Fig.~\ref{fig:delta-error} illustrates the variation of signed errors (in meters, left Y-axis) over the video timeline (seconds) for the different obstacle classes encountered. We also report the ground truth distances to the obstacles from the drone (purple line, right Y-axis). The regions shaded in gray background have the obstacle at $>4m$ from the drone, and hence are less crucial for VIP navigation. The regions with red background have the obstacle within $4m$ from the drone, and are critical for VIP navigation.

In Fig.~\ref{fig:temporalcar} for the \textit{car} class, both \neo-noRC and \neo-RC have a $\pm1.1m$ error in the red region of proximity while the error for Geometric is as high as $-3.0m$ and Geometric* is slightly better at $-2.0m$. Both the Geometric models overestimate the distances for all classes and time-points. A similar trend is seen for other classes, where both variants of \neo outperform. While the actual height of the car and scooter vary from the expected for Geometric, causing higher errors relative to Geometric*, they are comparable for cycle and bystanders and the blue lines overlap. \neo-RC performs similar to \neo-noRC, and is marginally better for cycle and scooter for true distances $\leq4m$. We we show next, this is even more prominent in adversarial scenes (\S~\ref{sec:evaluation:adv}) when \neo-RC recalibrates and adapts to dynamic scenarios.

Overall, we observe that both \neo variants consistently outperform the Geometric methods across all obstacle classes and time points, particularly in critical proximity regions where accurate distance estimation is vital for the VIP. 

%===================================================
\subsection{Robustness under Adversarial Scenarios}
\label{sec:evaluation:adv}

Finally, we perform a focused evaluation of the proposed models on adversarial scenarios for the dataset in \S~\ref{exp:data:adv} and shown in Fig.~\ref{fig:adversarialimages}. These challenging conditions cause the height of the YOLO bounding box to be incorrectly reduced for the VIP and bystander. Hence, the distance estimates for Regression and Geometric, which are sensitive to the bounding box accuracy, are adversely affected with large under-predictions of $60$--$90cm$ for VIP and $>120cm$ for bystander, as shown in Fig.~\ref{fig:adversarial}. However, \neo is able to predict distances with a median error of $\leq30cm$ in all cases. 

Here again, we report high errors for SOTA methods, with Monodepth2 showing a minimum error of $201cm$ while \midas+ZoeDepth reporting errors of $\geq119cm$ and going as high as $245cm$.

\subsection{Discussion and Limitations}\label{sec:discussion-limitations}
We summarize our observations of the benefits, limitations and overheads of the proposed methods in Table~\ref{fig:summary-fig}. For calibration and bootstrapping, Regression and \neo have a moderate overhead as they require manual collection of VIP images at fixed distances from the drone. Once calibrated, these do not require additional information. Geometric and Geo*, on the other hand, do not require any such calibration. But Geo* requires the exact height of any object, which is a costly or intractable, while Geo uses an average object height per class, incurring a moderate overhead.

\begin{table}
\centering
\renewcommand{\arraystretch}{1.2}
\setlength{\tabcolsep}{5.5pt}
\footnotesize
\caption{Comparison summary of different methods}
% \ysnote{cleanup header a bit, hanging 1st col} \srnote{Done.}
\label{fig:summary-fig}
% \begin{minipage}[b]{0.48\textwidth}
\begin{tabular}
% {|c|c|c|c|c|c|c|c|c|}
{|m{1.0cm}|m{1.15cm}|m{1.15cm}|m{1.15cm}|m{1.0cm}|m{1.0cm}|m{1.0cm}|m{1.0cm}|}
\hline
% \cline{2-8}
\multicolumn{1}{|c|}{\textbf{\textsc{Method}}} & \multicolumn{2}{c|}{\textbf{\textsc{Effort}}} & \multicolumn{4}{c|}{\textbf{\textsc{Error for Advers. Scenes}}} & 
\multirow{2}{1.6cm}
{\textbf{\textsc{Proc. Speed}}}\\ 
% \cline{2-7}
\cline{2-7}
\textbf{} & \em \textbf{Calib\-ration Overhead} & \em \textbf{Runtime Overhead} & \em \textbf{Varying Backgrounds} & \em \textbf{Orient\-ation} & \em \textbf{Diff. Indi\-viduals} & \em \textbf{Trunca\-ted Obj.} & \textbf{} \\ \hline
% \vline
\bf Regres.      & \cellcolor{yellow!50}Medium          
                & \cellcolor{green!35}Low                & \cellcolor{green!35}Low             
                & \cellcolor{orange!35}High       
                & \cellcolor{green!35}Low             
                & \cellcolor{orange!35}High              & \cellcolor{green!35}Low \\ \hline
% \vline
\bf Geo       & \cellcolor{green!35}Low             
                & \cellcolor{orange!35}Medium               & \cellcolor{green!35}Low             
                & \cellcolor{orange!35}High           
                & \cellcolor{green!35}Low            
                & \cellcolor{orange!35}High             
                & \cellcolor{green!35}Low \\ \hline
% \vline
\bf Geo*      & \cellcolor{green!35}Low             
                & \cellcolor{orange!35}High             & \cellcolor{green!35}Low             
                & \cellcolor{orange!35}High              
                & \cellcolor{green!35}Low            
                & \cellcolor{orange!35}High             
                & \cellcolor{green!35}Low \\ \hline
% \vline
\bf \neo       & \cellcolor{yellow!50}Medium          
                & \cellcolor{green!35}Low               & \cellcolor{yellow!50}Medium          
                & \cellcolor{green!35}Low              
                & \cellcolor{green!35}Low            
                & \cellcolor{green!35}Low               & \cellcolor{green!35}Low \\ \hline
\end{tabular}
% \end{minipage}
\end{table}

\neo generally performs well adversarial scenarios. In cases where the VIP changes their orientation by bending or sitting (Fig.~\ref{fig:adversarial}), \neo has low errors of distance estimation, whereas the other methods fail due of their dependency on the bounding box parameters (height/width/area). Similarly, when objects are truncated due to their closeness to the camera, \neo provides better results. However, the distance estimation of \neo is affected by the background and lighting conditions. 
All methods generalize well across individuals. So, a model trained on one VIP works well on any other VIP. Finally, as reported in Table~\ref{table:dnn-config}, all methods take $<50ms$ of processing time per frame on the Jetson Orin Nano edge accelerator.

There are limitations for \neo as well, which are worth addressing as future work for robust deployment. Its accuracy is highly reliant on the bounding box contents for the VIP and its heatmap. Hence, it is crucial to use a high quality detection model. Transient conditions such as motion blur or image blur should also be detected and corrected, often by skipping frames.
During recalibration, where \neo uses Regression as the ground truth, scenarios such as the VIP bending or falling can cause Regression to have estimation errors. These frames should be treated as outliers and skipped during recalibration to prevent incorrect updates and maintain robustness. Such scenes can be detected using Body Pose Classification models. The framework is also restricted by some operational challenges. Small consumer drones have limited battery capacity, currently ranging in 10s of minutes. Lightweight drones may fail in adverse environmental conditions like heavy rains, wind gusts, etc. Further, our design assumes that the heading of the VIP and drone are in the same direction to simplify the analysis. However, when tracking VIPs in real-life scenarios, the drone's orientation be different from the VIP's. Models to dynamically estimate the VIP's orientation and relative heading of the drone can be used to correct the model's distance estimates.

%% #############################################################################################
\section{Related Work}\label{sec:related-work}

\subsubsection*{Vision-based Real-Time Obstacle Avoidance}
Algorithms over images and videos have been developed for collision avoidance. Early works such as Watanabe, et al.~\cite{watanabe2007vision} use a single 2D passive vision sensor and an extended Kalman filter to estimate the relative position of obstacles using a collision-cone approach. Recent advances in machine perception and GPU-accelerated computing have led to a large body of literature that integrates computer vision-based approaches with deep learning~\cite{5651028}. Here, obstacle avoidance is often vision-based, exploiting DNNs to extract the navigation context from the scene~\cite{loquercio2021learning}. We leverage some of these advances in our work.

Kaff et al.~\cite{al2017obstacle} propose an obstacle detection algorithm for UAVs equipped with monocular cameras, which analyzes changes in the pixel area of approaching obstacles. Obstacles are avoided by estimating their 2D position in the image and combining it with tracked waypoints. Similarly, Lee et al.~\cite{lee2021deep} utilize DNN models for monocular obstacle avoidance in UAVs flying through tree plantations, classifying obstacles as critical or low priority based on the height of the bounding boxes around detected trees. However, these approaches are not designed to estimate absolute distances to objects, a crucial requirement for assisting VIPs in navigation. Our current work focuses on reliably determining the distance to each object in the frame to help classify them as obstacles, and later plan a collision-free path as future work.

%=================================================
\subsubsection*{Depth-Map Based Navigation Techniques}
Depth-map based navigation techniques are widely used in autonomous systems for spatial awareness and obstacle detection~\cite{muller2023robust}. Estimating a scene's depth is crucial, and integrating it with deep learning has shown success in achieving collision avoidance against moving obstacles and dynamic environments~\cite{liu2019deepnav}. While stereo vision provides robust depth measurements under diverse lighting conditions and dynamic scenarios~\cite{zhang2020stereodepth}, monocular depth estimation has gained popularity as a cost-effective alternative due to its reliance on a single camera~\cite{fu2018deep}. However, monocular approaches often struggle with accurately predicting absolute distances, relying heavily on relative depth cues~\cite{eigen2014depth,monodepth2}. Additionally, the quality and accuracy of depth maps are highly sensitive to lighting conditions, with poor illumination or sharp shadows significantly degrading depth estimation performance. Unlike these, \neo focuses on estimating absolute distances using lightweight calibration techniques that dynamically recalibrate depth coefficients in response to environmental variations. This ensures both adaptability and reliability for real-world navigation.

Accurately estimating absolute distances from monocular views remains a core challenge without additional sensory inputs, such as stereo or LiDAR~\cite{fu2018deep}. The Depth visUal Ego-motion Learning (DUEL) model combines depth perception with ego-motion estimation to enable autonomous robots to effectively avoid obstacles~\cite{wang2024duel}. However, its high compute demand limits practical deployment. Edge devices are being leveraged for depth map based navigation in UAVs and ground robots. Such local processing minimizes latency and enhances real-time navigation~\cite{sampedro2018uavdepth}. An et al.~\cite{9636518} propose a low-complexity network optimized for fast human depth estimation and segmentation in indoor environments, tailored for resource-constrained edge devices. Building on these efforts, \neo focuses on outdoor applications and leverages Jetson edge accelerators for dynamic recalibration while achieving real-time performance. 

%=================================================
\subsubsection*{Absolute Distance Estimation Methods}
Absolute distance estimation is a critical component in robotics and autonomous navigation. Traditional methods, such as stereo vision and LiDAR, directly measure distances based on geometric triangulation or light reflection times, providing high accuracy in estimating absolute distances. But they require expensive hardware and significant processing power~\cite{szeliski2022computer}. SOTA methods for distance estimation integrate camera feeds with LiDAR~\cite{8793729} or use RGB-D cameras~\cite{iacono2018path} to generate a 3D view of the surroundings. We limit ourselves to just monocular images from the camera present on affordable consumer drones for wider applicability. 

In monocular vision systems, there is an \textit{inherent scale ambiguity} where the actual size and distance of objects cannot be distinguished. This makes it challenging to gauge the true depth without assumptions about object size or additional sensors for distance verification~\cite{mur2017orb}. We observe a similar issue while evaluating our baseline Geometric method that estimates the distance to an object using only the average height of the object class. We see that the estimation errors increase when the VIP is partially out of the frame or if the object size estimations deviate from the actual sizes. \neo's calibration techniques adapt well to such variations.

Deep reinforcement learning using monocular vision~\cite{liu2019deepnav} and end-to-end learning have also been implemented for autonomous vehicles~\cite{bojarski2016end}, but require massive training datasets. Some use geometric methods~\cite{joglekar2011depth} on static objects but need details on the road geometry and point of contact on the road. We instead focus on lightweight methods that operate on monocular images with minimal ancillary metadata, and attempt to generalize to any object seen by the UAV.

%% #############################################################################################
%% #############################################################################################
\section{Conclusion and Future Work}\label{sec:conclusions}
In this paper, we proposed a novel approach, NeoARCADE, which leverages depth maps to estimate absolute distances to obstacles in outdoor urban environments, facilitating obstacle detection and eventually autonomous navigation for VIPs. \neo uses robust calibration methods to address the limitations of SOTA depth map approaches like \midas and MonoDepth2, which offer poor accuracy for real distances.  Additionally, we present Geometric and Regression-based methods along with their calibration techniques as baselines. We also design dynamic recalibration methods for \neo to adapt to changing scenes.
We evaluate their efficacy on diverse videos at various distances, backgrounds and complexs environments in outdoor campus settings, captured using a Tello drone. While Regression calibrated on specific VIPs works best for the VIP's distance estimate, \neo performs consistently better for all other obstacles, scales well to dynamic changes, and outperforms in adversarial conditions. It also does much better than the SOTA methods.

As future work, we plan to integrate these with path-planning algorithms for the local navigation of the VIPs. Also, further investigation is needed to enhance the accuracy of depth map methods, which can be explored in future work. Additionally, we plan to extend our validation to other complex scenarios with dynamic obstacles emulating real-time use-cases to provide enhanced collision avoidance guidance to VIPs and the drone.

%% #############################################################################################
\section*{Acknowledgments}
The authors would like to thank Prof. Debasish Ghose, Arnav Rajesh, Pratham M, Ansh Bhatia, Prince Modi, Rishubh Parihar, Dibyajyoti Nayak, Pranjal Naman, Gourab Panigrahi, Thanushree R, Lakshya, Beautlin, Radhika Mittal, Varad Kulkarni and Suved Ghanmode from IISc for their assistance.  The first author was supported by Prime Minister's Research Fellowship (PMRF) from the Government of India.

\balance
\bibliographystyle{unsrt}
\bibliography{references}

\begin{thebibliography}{10}

\bibitem{drone-first-aid}
Alec Momont.
\newblock \href{https://www.tudelft.nl/en/ide/research/research-labs/applied-labs/ambulance-drone/}{Ambulance Drone at TU Delft}.
\newblock Technical report, TUDelft, 2014.

\bibitem{ollero2004motion}
An{\'\i}bal Ollero, Joaquin Ferruz, Fernando Caballero, Sebasti{\'a}n Hurtado, and Luis Merino.
\newblock Motion compensation and object detection for autonomous helicopter visual navigation in the comets system.
\newblock In {\em IEEE International Conference on Robotics and Automation (ICRA)}, volume~1, pages 19--24, New Orleans, Louisiana, USA, 2004. IEEE.

\bibitem{10.1145/3450356}
Stephanie Abrecht, Lydia Gauerhof, Christoph Gladisch, Konrad Groh, Christian Heinzemann, and Matthias Woehrle.
\newblock Testing deep learning-based visual perception for automated driving.
\newblock {\em ACM Tran. Cyber-Phys. Syst.}, 5(4), 2021.

\bibitem{arafat2023vision}
Muhammad~Yeasir Arafat, Muhammad~Morshed Alam, and Sangman Moh.
\newblock Vision-based navigation techniques for unmanned aerial vehicles: Review and challenges.
\newblock {\em Drones}, 7(2):89, 2023.

\bibitem{10171496}
Suman Raj, Harshil Gupta, and Yogesh Simmhan.
\newblock Scheduling dnn inferencing on edge and cloud for personalized uav fleets.
\newblock In {\em 2023 IEEE/ACM 23rd International Symposium on Cluster, Cloud and Internet Computing (CCGrid)}, pages 615--626, Bangalore, India, 2023. IEEE/ACM.

\bibitem{RAJ2025107874}
Suman Raj, Radhika Mittal, Harshil Gupta, and Yogesh Simmhan.
\newblock Adaptive heuristics for scheduling dnn inferencing on edge and cloud for personalized uav fleets.
\newblock {\em Future Generation Computer Systems}, 173:107874, 2025.

\bibitem{al2016exploring}
Majed Al~Zayer, Sam Tregillus, Jiwan Bhandari, Dave Feil-Seifer, and Eelke Folmer.
\newblock Exploring the use of a drone to guide blind runners.
\newblock In {\em ACM SIGACCESS}, ASSETS, pages 263--264, New York, NY, USA, 2016. ACM.

\bibitem{whoStats}
World~Health Organization.
\newblock \href{https://www.who.int/news-room/fact-sheets/detail/blindness-and-visual-impairment}{Blindness and visual impairment fact sheets}.
\newblock Technical report, WHO, August 2023.

\bibitem{suman2023chi}
Suman Raj, Swapnil Padhi, and Yogesh Simmhan.
\newblock Ocularone: Exploring drones-based assistive technologies for the visually impaired.
\newblock In {\em Extended Abstracts of the CHI Conference on Human Factors in Computing Systems}. ACM, 2023.

\bibitem{wewalk}
{WeWALK Limited UK}.
\newblock \href{https://wewalk.io/en/}{WeWalk: Enhancing the mobility of visually impaired people}, August 2020.

\bibitem{bai2017smart}
Jinqiang Bai, Shiguo Lian, Zhaoxiang Liu, Kai Wang, and Dijun Liu.
\newblock Smart guiding glasses for visually impaired people in indoor environment.
\newblock {\em IEEE Transactions on Consumer Electronics}, 63(3):258--266, 2017.

\bibitem{lookout}
Abrar Al-Heeti.
\newblock Google expands lookout app for people who are blind or vision-impaired.
\newblock Technical report, CNet, August 2020.

\bibitem{nasralla2019computer}
Moustafa~M. Nasralla, Ikram~U. Rehman, Drishty Sobnath, and Sara Paiva.
\newblock Computer vision and deep learning-enabled uavs: Proposed use cases for visually impaired people in a smart city.
\newblock In {\em Computer Analysis of Images and Patterns}, pages 91--99, Salerno, Italy, 2019. Springer International Publishing.

\bibitem{avila2017dronenavigator}
Mauro Avila~Soto, Markus Funk, Matthias Hoppe, Robin Boldt, Katrin Wolf, and Niels Henze.
\newblock Dronenavigator: Using leashed and free-floating quadcopters to navigate visually impaired travelers.
\newblock In {\em International ACM SIGACCESS Conference on Computers and Accessibility}, ASSETS '17, 2017.

\bibitem{10802577}
Haokun Zheng, Sidhant Rajadnya, and Avideh Zakhor.
\newblock Monocular depth estimation for drone obstacle avoidance in indoor environments.
\newblock In {\em IEEE/RSJ International Conference on Intelligent Robots and Systems (IROS)}, 2024.

\bibitem{scharstein2002taxonomy}
Daniel Scharstein and Richard Szeliski.
\newblock A taxonomy and evaluation of dense two-frame stereo correspondence algorithms.
\newblock {\em International journal of computer vision}, 47:7--42, 2002.

\bibitem{zhang2012microsoft}
Zhengyou Zhang.
\newblock Microsoft kinect sensor and its effect.
\newblock {\em IEEE multimedia}, 19(2):4--10, 2012.

\bibitem{mur2017orb}
Raul Mur-Artal and Juan~D Tard{\'o}s.
\newblock Orb-slam2: An open-source slam system for monocular, stereo, and rgb-d cameras.
\newblock {\em IEEE Transactions on Robotics}, 33(5):1255--1262, 2017.

\bibitem{ranftl2020towards}
Ren{\'e} Ranftl, Katrin Lasinger, David Hafner, Konrad Schindler, and Vladlen Koltun.
\newblock Towards robust monocular depth estimation: Mixing datasets for zero-shot cross-dataset transfer.
\newblock {\em IEEE transactions on pattern analysis and machine intelligence}, 44(3):1623--1637, 2020.

\bibitem{monodepth2}
Cl{\'{e}}ment Godard, Oisin {Mac Aodha}, Michael Firman, and Gabriel~J. Brostow.
\newblock Digging into self-supervised monocular depth prediction.
\newblock In {\em The International Conference on Computer Vision (ICCV)}, pages 3828--3838, Seoul, Korea (South), October 2019. IEEE.

\bibitem{zoedepth}
Shariq~Farooq Bhat, Reiner Birkl, Diana Wofk, Peter Wonka, and Matthias Müller.
\newblock Zoedepth: Zero-shot transfer by combining relative and metric depth, 2023.

\bibitem{ranftl2021vision}
Rene Ranftl, Alexey Bochkovskiy, and Vladlen Koltun.
\newblock Vision transformers for dense prediction.
\newblock {\em IEEE International Conference on Computer Vision (ICCV)}, pages 12179--12188, 2021.

\bibitem{al2017obstacle}
Abdulla Al-Kaff, Fernando Garc{\'\i}a, David Mart{\'\i}n, Arturo De~La~Escalera, and Jos{\'e}~Mar{\'\i}a Armingol.
\newblock Obstacle detection and avoidance system based on monocular camera and size expansion algorithm for uavs.
\newblock {\em Sensors}, 17(5), 2017.

\bibitem{lee2021deep}
HY~Lee, Hann~Woei Ho, and Ye~Zhou.
\newblock Deep learning-based monocular obstacle avoidance for unmanned aerial vehicle navigation in tree plantations: Faster region-based convolutional neural network approach.
\newblock {\em Journal of Intelligent \& Robotic Systems}, 101(5):5, 2021.

\bibitem{wang2024duel}
Naiyao Wang, Bo~Zhang, Haixu Chi, Hua Wang, Se{\'a}n McLoone, and Hongbo Liu.
\newblock Duel: Depth visual ego-motion learning for autonomous robot obstacle avoidance.
\newblock {\em The International Journal of Robotics Research}, 43(3), 2024.

\bibitem{10610042}
Xiongfeng Peng, Zhihua Liu, Weiming Li, Ping Tan, Soon~Yong Cho, and Qiang Wang.
\newblock Dvi-slam: A dual visual inertial slam network.
\newblock In {\em International Conference on Robotics and Automation (ICRA)}. IEEE, 2024.

\bibitem{kitti_dataset}
Andreas Geiger, Philipp Lenz, and Raquel Urtasun.
\newblock Kitti vision benchmark suite: Raw data.
\newblock \url{https://www.cvlibs.net/datasets/kitti/raw_data.php}, 2012.
\newblock Accessed: 2025-01-05.

\bibitem{sumaniros}
Suman Raj, Swapnil Padhi, Ruchi Bhoot, Prince Modi, and Yogesh Simmhan.
\newblock Towards perception-based collision avoidance for uavs when guiding the visually impaired.
\newblock 2025.

\bibitem{cloutier2022topical}
Melissa Cloutier and Patricia~R DeLucia.
\newblock Topical review: Impact of central vision loss on navigation and obstacle avoidance while walking.
\newblock {\em Optometry and Vision Science}, 99(12):890--899, 2022.

\bibitem{yolov8_ultralytics}
Glenn Jocher, Ayush Chaurasia, and Jing Qiu.
\newblock \href{https://github.com/ultralytics/ultralytics}{Ultralytics YOLOv8}, 2023.

\bibitem{Sturm2014}
Peter Sturm.
\newblock Pinhole camera model.
\newblock In {\em Computer Vision: A Reference Guide}, pages 610--613. Springer US, Boston, MA, 2014.

\bibitem{Hartley2004}
R.~I. Hartley and A.~Zisserman.
\newblock {\em Multiple View Geometry in Computer Vision}.
\newblock Cambridge University Press, Cambridge, England, second edition, 2004.

\bibitem{eigen2014depth}
David Eigen, Christian Puhrsch, and Rob Fergus.
\newblock Depth map prediction from a single image using a multi-scale deep network.
\newblock In {\em International Conference on Neural Information Processing Systems}, 2014.

\bibitem{bhat2021adabins}
Shariq Farooq~Bhat, Ibraheem Alhashim, and Peter Wonka.
\newblock Adabins: Depth estimation using adaptive bins.
\newblock In {\em 2021 IEEE/CVF Conference on Computer Vision and Pattern Recognition (CVPR)}, pages 4008--4017, Nashville, TN, USA, 2021. IEEE.

\bibitem{yu2023udepth}
Boxiao Yu, Jiayi Wu, and Md~Jahidul Islam.
\newblock Udepth: Fast monocular depth estimation for visually-guided underwater robots.
\newblock In {\em International Conference on Robotics and Automation (ICRA)}, pages 3116--3123. IEEE, 2023.

\bibitem{ke2023repurposing}
Bingxin Ke, Anton Obukhov, Shengyu Huang, Nando Metzger, Rodrigo~Caye Daudt, and Konrad Schindler.
\newblock Repurposing diffusion-based image generators for monocular depth estimation.
\newblock In {\em Proceedings of the IEEE/CVF Conference on Computer Vision and Pattern Recognition (CVPR)}, pages 9492--9502, Seattle, WA, USA, 2024. IEEE.

\bibitem{Alhashim2018}
Ibraheem Alhashim and Peter Wonka.
\newblock High quality monocular depth estimation via transfer learning, 2019.

\bibitem{tello}
Ryze Tech.
\newblock \href{https://www.ryzerobotics.com/tello}{RYZE Tello: Powered by DJI}, 2018.

\bibitem{jetsonorinnano}
NVIDIA Developer.
\newblock \href{https://developer.nvidia.com/embedded/learn/get-started-jetson-orin-nano-devkit}{Jetson Orin Nano Developer Kit}, 2023.

\bibitem{raj2025ocularonebenchbenchmarkingdnnmodels}
Suman Raj, Bhavani~A Madhabhavi, Kautuk Astu, Arnav~A Rajesh, Pratham M, and Yogesh Simmhan.
\newblock Ocularone-bench: Benchmarking dnn models on gpus to assist the visually impaired.
\newblock In {\em 2025 IEEE International Parallel and Distributed Processing Symposium Workshops (IPDPSW)}, pages 251--254, 2025.

\bibitem{watanabe2007vision}
Yoko Watanabe, Anthony Calise, and Eric Johnson.
\newblock Vision-based obstacle avoidance for uavs.
\newblock In {\em AIAA guidance, navigation and control conference and exhibit}, page 6829, 2007.

\bibitem{5651028}
Luis Mejias, Scott McNamara, John Lai, and Jason Ford.
\newblock Vision-based detection and tracking of aerial targets for uav collision avoidance.
\newblock In {\em IEEE/RSJ International Conference on Intelligent Robots and Systems}, pages 87--92, Taipei, Taiwan, 2010. IEEE.

\bibitem{loquercio2021learning}
Antonio Loquercio, Elia Kaufmann, Ren{\'e} Ranftl, Matthias M{\"u}ller, Vladlen Koltun, and Davide Scaramuzza.
\newblock Learning high-speed flight in the wild.
\newblock {\em Science Robotics}, 6(59):eabg5810, 2021.

\bibitem{muller2023robust}
Hanna M{\"u}ller, Vlad Niculescu, Tommaso Polonelli, Michele Magno, and Luca Benini.
\newblock Robust and efficient depth-based obstacle avoidance for autonomous miniaturized uavs.
\newblock {\em IEEE Transactions on Robotics}, 39(6):4935--4951, 2023.

\bibitem{liu2019deepnav}
Xiang Liu, Zhiyong Zhu, and Hao Zhang.
\newblock Deep reinforcement learning for navigation of uavs with obstacle avoidance.
\newblock {\em IEEE Access}, 7:156720--156732, 2019.

\bibitem{zhang2020stereodepth}
Jiayi Zhang, Xin Wang, and Lei Xie.
\newblock Stereo depth estimation network for real-time stereo vision applications.
\newblock {\em IEEE Transactions on Industrial Electronics}, 67(11):9392--9402, 2020.

\bibitem{fu2018deep}
Huan Fu, Mingming Gong, Chaohui Wang, Kayhan Batmanghelich, and Dacheng Tao.
\newblock Deep ordinal regression network for monocular depth estimation.
\newblock In {\em Proceedings of the IEEE conference on computer vision and pattern recognition}, pages 2002--2011, Salt Lake City, UT, USA, 2018. IEEE.

\bibitem{sampedro2018uavdepth}
Carlos Sampedro, Harish Bavle, Andres Carrio, and Pascual Campoy.
\newblock Uav autonomous navigation and obstacle avoidance using depth maps and artificial potential fields.
\newblock {\em IEEE Robotics and Automation Letters}, 3(4):3271--3278, 2018.

\bibitem{9636518}
Shan An, Fangru Zhou, Mei Yang, Haogang Zhu, Changhong Fu, and Konstantinos~A Tsintotas.
\newblock Real-time monocular human depth estimation and segmentation on embedded systems.
\newblock In {\em IEEE/RSJ IROS}, 2021.

\bibitem{szeliski2022computer}
Richard Szeliski.
\newblock {\em Computer vision: algorithms and applications}.
\newblock Springer Nature, 2022.

\bibitem{8793729}
Zimo Li, Prakruti~C. Gogia, and Michael Kaess.
\newblock Dense surface reconstruction from monocular vision and lidar.
\newblock In {\em ICRA}, pages 6905--6911, Montreal, QC, Canada, 2019. IEEE.

\bibitem{iacono2018path}
Massimiliano Iacono and Antonio Sgorbissa.
\newblock Path following and obstacle avoidance for an autonomous uav using a depth camera.
\newblock {\em Robotics and Autonomous Systems}, 106:38--46, 2018.

\bibitem{bojarski2016end}
Mariusz Bojarski, Davide~Del Testa, Daniel Dworakowski, Bernhard Firner, Beat Flepp, Prasoon Goyal, Lawrence~D. Jackel, Mathew Monfort, Urs Muller, Jiakai Zhang, Xin Zhang, Jake Zhao, and Karol Zieba.
\newblock End to end learning for self-driving cars, 2016.

\bibitem{joglekar2011depth}
Apoorva Joglekar, Devika Joshi, Richa Khemani, Smita Nair, and Shashikant Sahare.
\newblock Depth estimation using monocular camera.
\newblock {\em International journal of computer science and information technologies}, 2(4):1758--1763, 2011.

\end{thebibliography}

\end{document}